\documentclass[sigconf]{acmart}

\AtBeginDocument{%
  \providecommand\BibTeX{{%
    \normalfont B\kern-0.5em{\scshape i\kern-0.25em b}\kern-0.8em\TeX}}}

\copyrightyear{2021}
\acmYear{2021}
\setcopyright{acmcopyright}\acmConference[CIKM '21]{Proceedings of the 30th ACM International Conference on Information and Knowledge Management}{November 1--5, 2021}{Virtual Event, QLD, Australia}
\acmBooktitle{Proceedings of the 30th ACM International Conference on Information and Knowledge Management (CIKM '21), November 1--5, 2021, Virtual Event, QLD, Australia}
\acmPrice{15.00}
\acmDOI{10.1145/3459637.3482471}
\acmISBN{978-1-4503-8446-9/21/11}

\graphicspath{ {images/} }

\usepackage{multirow}
\usepackage{subfigure}
\usepackage{amsfonts}
\usepackage[T1]{fontenc}
\usepackage{aecompl}
\usepackage{amsmath}

\begin{document}

\fancyhead{}

\title{Region Semantically Aligned Network for Zero-Shot Learning}

\author{Ziyang Wang}
\authornote{Equal contribution, authorship order determined by coin flipping.}
\email{tedwzy2000@gmail.com}
\affiliation{%
  \institution{University of Electronic Science and Technology of China}
  \city{Chengdu}
  \country{China}
}

\author{Yunhao Gou}
\authornotemark[1]
\email{yhaogou@gamil.com}
\affiliation{%
  \institution{University of Electronic Science and Technology of China}
  \city{Chengdu}
  \country{China}
}

\author{Jingjing Li}
\authornote{Corresponding author.}
\email{lijin117@yeah.net}
\affiliation{%
  \institution{University of Electronic Science and Technology of China}
  \city{Chengdu}
  \country{China}
}

\author{Yu Zhang}
\email{yu.zhang.ust@gmail.com}
\affiliation{%
  \institution{Southern University of Science and Technology}
  \city{Shenzhen}
  \country{China}
}

\author{Yang Yang}
\email{dlyyang@gmail.com}
\affiliation{%
  \institution{University of Electronic Science and Technology of China}
  \city{Chengdu}
  \country{China}
}


\begin{abstract}
Zero-shot learning (ZSL) aims to recognize unseen classes based on the knowledge of seen classes. Previous methods focused on learning direct embeddings from global features to the semantic space in hope of knowledge transfer from seen classes to unseen classes. However, an unseen class shares local visual features with a set of seen classes and leveraging global visual features makes the knowledge transfer ineffective. To tackle this problem, we propose a Region Semantically Aligned Network (RSAN), which maps local features of unseen classes to their semantic attributes. Instead of using global features which are obtained by an average pooling layer after an image encoder, we directly utilize the output of the image encoder which maintains local information of the image. Concretely, we obtain each attribute from a specific region of the output and exploit these attributes for recognition. As a result, the knowledge of seen classes can be successfully transferred to unseen classes in a region-bases manner. In addition, we regularize the image encoder through attribute regression with a semantic knowledge to extract robust and attribute-related visual features. Experiments on several standard ZSL datasets reveal the benefit of the proposed RSAN method, outperforming state-of-the-art methods.

\end{abstract}




\maketitle

\begin{figure}
\centering
\includegraphics[scale=0.35]{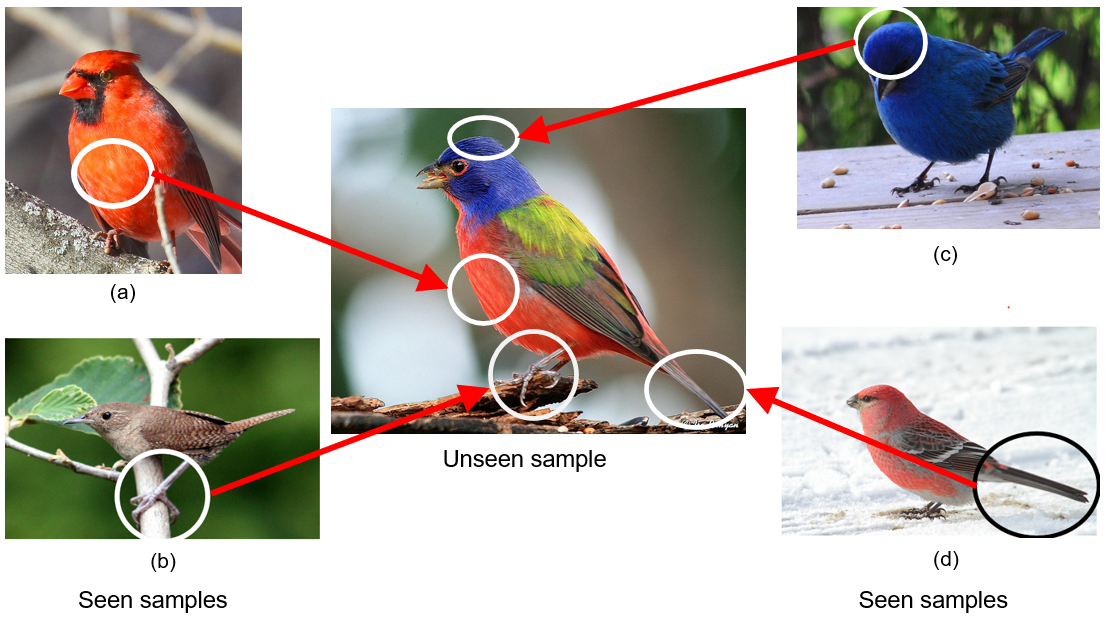}
\caption{Motivation of this paper. The seen and unseen samples relate to each other by similar image regions as illustrated. The unseen sample shares orange belly, leg pattern, blue crown and tail pattern with seen sample (a), (b), (c), (d). Our work directly maps the image regions of unseen sample to its semantic attributes through the knowledge from the image regions of seen samples. }
\label{fig:Intro}
\end{figure}

\section{Introduction}
Deep learning has accomplished a great success in various tasks, especially in supervised image recognition \cite{Krizhevsky2012ImageNetCW}. However, supervised learning requires a large quantity of labeled data which are expensive to obtain and even unavailable in a real world scenario. Zero-shot learning (ZSL) \cite{Lampert2009LearningTD, Palatucci2009ZeroshotLW, Larochelle2008ZerodataLO, Lampert2014AttributeBasedCF, Pourpanah2020ARO}, which aims to recognize unseen classes through the knowledge from seen classes, has shown the potential to avoid collecting large-scale labeled samples. To tackle the absence of unseen samples, ZSL methods exploit semantic descriptions \cite{Lampert2009LearningTD, Farhadi2009DescribingOB, Socher2013ZeroShotLT, Fu2018ZeroShotLO} which describe the characteristics of both seen and unseen classes.  Most semantic descriptions are composed of a group of high-level class attributes \cite{Lampert2009LearningTD, Farhadi2009DescribingOB}, such as shape (e.g., round), color (e.g., red), pattern (e.g., striped), which not only depict the class objects, but also connect the unseen classes to the seen classes. The knowledge of semantic attributes are learned from the seen classes and then transferred to the unseen classes for recognition.

Previous methods \cite{Akata2013LabelEmbeddingFA, Akata2015EvaluationOO, Frome2013DeViSEAD, Morgado2017SemanticallyCR, RomeraParedes2015AnES} on ZSL established an embedding between visual space and semantic space via seen samples and their semantic descriptions. Then unseen samples are recognized by the predicted semantic representation of the image through this embedding. Specifically, \cite{Frome2013DeViSEAD, Akata2015EvaluationOO, RomeraParedes2015AnES, Zhang2015ZeroShotLV, Fu2018ZeroShotLO} mapped the visual features to semantic space while \cite{Shigeto2015RidgeRH, Das2019ZeroshotIR, Zhang2019CoRepresentationNF} mapped the semantic representations to visual space. To alleviate the hubness problem \cite{Arora2018GeneralizedZL, Zhang2017LearningAD}, \cite{Zhang2017LearningAD, Yang2018DissimilarityRL} leveraged a latent space as an intermediary between visual and semantic space. With the help of generative models \cite{Goodfellow2014GenerativeAN, Kingma2014AutoEncodingVB}, \cite{Xian2018FeatureGN, Zhu2018AGA, Changpinyo2017PredictingVE, Li2019LeveragingTI, Li2019AlleviatingFC} generated unseen samples via their semantic descriptions and convert a ZSL problem to a supervised learning problem. As the embeddings or generators from ZSL methods are learned only from seen samples, they inevitably bias to seen classes in Generalized Zero-shot Learning settings (GZSL) \cite{Chao2016AnES} where the test samples are comprised of both seen and unseen samples. To tackle this problem, \cite{Vyas2020LeveragingSA} considered the semantic relationships between unseen classes and seen classes, and \cite{Chen2020ABB, Yue2021CounterfactualZA} proposed the gate method to discriminate seen and unseen domains to convert a GZSL problem to a ZSL problem plus a supervised learning problem.  

It is worth noting that most of the existing embedding and feature generation methods extract global features from an end-to-end or pre-trained network \cite{RomeraParedes2015AnES, Changpinyo2016SynthesizedCF, Akata2016LabelEmbeddingFI, Zhang2017LearningAD}. However, the global features lack fine-grained information of the image, which is essential to the knowledge transfer in ZSL. As shown in Figure \ref{fig:Intro}, an unseen sample shares different partial information with a set of seen samples and this part information represents the knowledge of semantic attributes. Since each of the shared parts only takes up a small area of the unseen sample, global features of the unseen sample fail to represent those part information and lead to a negative effect on knowledge transfer from seen classes to unseen classes. Recently, several methods \cite{Huynh2020FineGrainedGZ, Xu2020AttributePN, Liu2021GoalOrientedGE, Xie2020RegionGE} focused on utilizing part information in ZSL. Huynh et al. \cite{Huynh2020FineGrainedGZ} proposed a dense attention mechanism that for each attribute focuses on the relevant image regions. Xie et al. \cite{Xie2020RegionGE} leveraged a region graph to accomplish region-based relation reasoning. Xu et al. \cite{Xu2020AttributePN} jointly learned globally and locally discriminative features for knowledge transfer. Liu et al. \cite{Liu2021GoalOrientedGE} leveraged a gaze estimation module to predict the actual human gaze location to get the visual attention regions for recognition. However, the local features in \cite{Xu2020AttributePN,Liu2021GoalOrientedGE} are only used as supplements of global features instead of being directly utilized for recognition when facing an unseen sample. To reserve every detail information, in our work, rather than adopting the commonly-used average pooling layer to extract global features, we directly exploit the outputs of the image encoder as our visual features. 

Interestingly, when facing an unseen sample, humans tend to scrutinize the sample and find the regions which are similar to the regions of seen samples \cite{Meacham1983WisdomAT}. The knowledge of seen classes is transferred to unseen samples in a region-based manner in human brains. Inspired by this, we propose a Region Semantically Aligned Network (RSAN) for Zero-Shot Learning. Our goal is to map the visual features of unseen image regions to their semantic attributes with the knowledge of seen classes. However, as we only have the image-wise annotations, our work can be considered as a weakly supervised learning model and it is difficult to localize the attributes. To handle this problem, we first partition the image to a set of image regions. Then we respectively compute the confidence of each region to have a specific semantic attribute and get a possibility map of each attribute. Through the confidence map, we are able to localize each semantic attributes and get attribute representations from the corresponding regions. By integrating all attribute representations, we obtain the semantic representations of the image for recognition. In addition, to benefit the process of attribute localization, we leverage attribute regression to enforce the image encoder to extract visual features which are related to semantic attributes. However, the domain shift problem \cite{Jia2020DeepUE, FuYanwei2015TransductiveMZ, Zhao2017ZeroshotLV} impedes the image encoder to learn valid attribute patterns for unseen sample. To mitigate this problem, we further utilize a semantic knowledge as auxiliary information.

To sum up, our contributions are as follows:

(1) We directly utilize the output of image encoder to reserve the valuable detail information of the samples. We further exploit this information as the knowledge which can be transferred from seen classes to unseen classes.

(2) We propose a Region Semantically Aligned Network (RSAN) which builds a direct connection between the visual features of image regions and semantic attributes. With the help of this connection, the knowledge from seen classes can be successfully transferred to unseen classes for recognition. We further enforce the image encoder to learn visual features that are related to semantic attributes to improve the localization of attributes.

(3) We evaluate the proposed RSAN on three benchmark datasets and report state-of-the-art or competitive results under both ZSL and GZSL settings. Moreover, the proposed model achieves a significant improvement on those classes which are hard to discriminate compared with previous methods.

\begin{figure*}[t]
\centering

\includegraphics[width=0.8\textwidth]{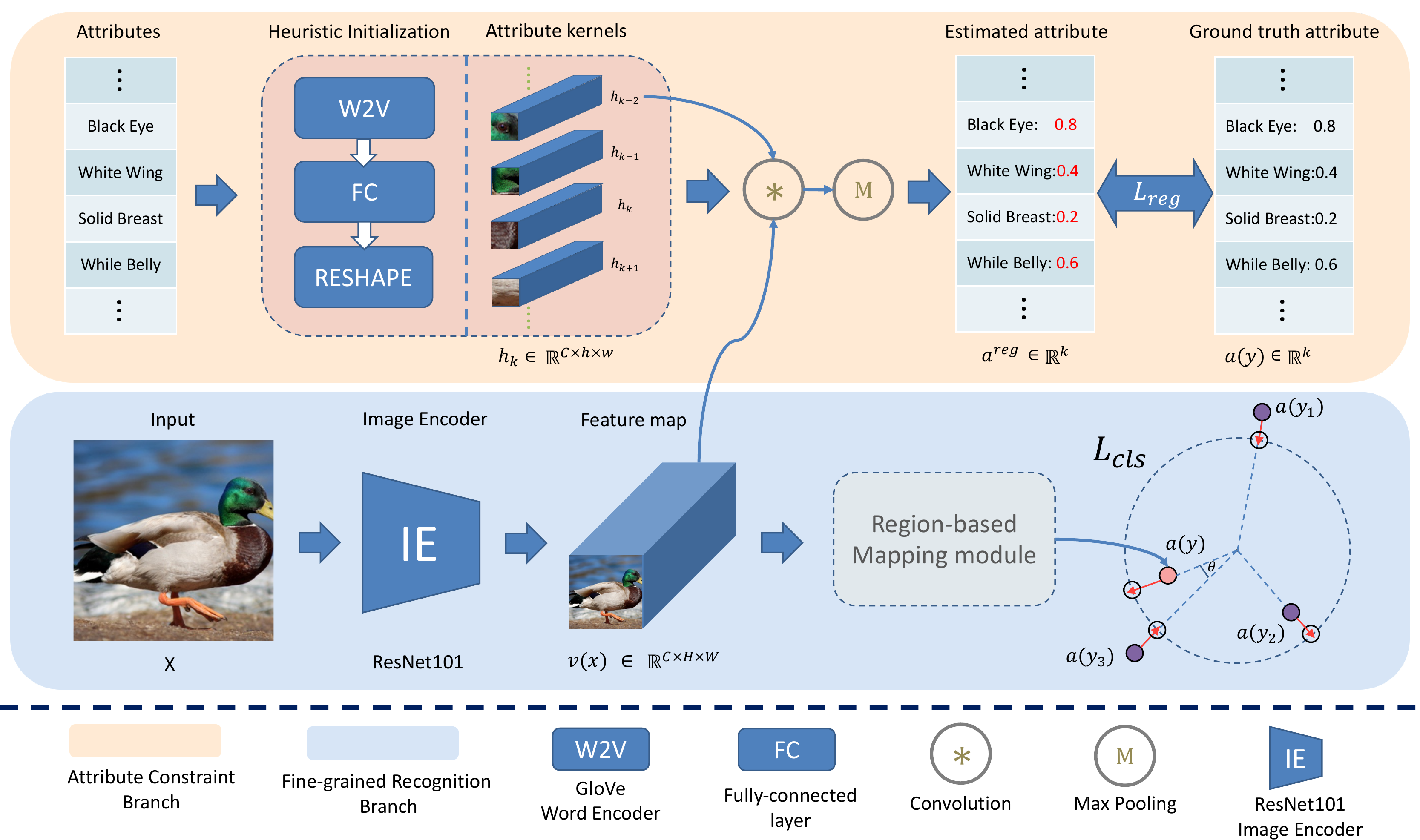}

\caption{Illustration of the RSAN framework. Fine-grained Recognition branch (FR) and Attribute Constraint branch (AC) share the Image Encoder. 
}
\label{fig:framwork}

\end{figure*}
\section{Related work}
\textbf{Zero-shot learning.}
ZSL \cite{Lampert2009LearningTD, Palatucci2009ZeroshotLW, Larochelle2008ZerodataLO, Lampert2014AttributeBasedCF} aims to transfer the knowledge from seen classes to unseen classes via semantic descriptions. Most existing ZSL methods can be divided into two categories:  embedding-based methods \cite{Akata2013LabelEmbeddingFA, Frome2013DeViSEAD, Morgado2017SemanticallyCR, RomeraParedes2015AnES} and generative-based methods \cite{Xian2018FeatureGN, Zhu2018AGA,Li2019FromZL,Li2021InvestigatingTB, Li2019LeveragingTI, Li2019AlleviatingFC, Chen2020CANZSLCA, Chen2020RethinkingGZ,Li2020LearningML}. Generative-based methods generate samples for unseen classes with the guidance of their semantic descriptions and convert the ZSL problem to a supervised learning problem. However, it is difficult to synthesize discriminative data samples from semantic descriptions, due to the overlap of common features such as color and shape between many classes. The generated unseen samples can be easily mistaken as the similar seen classes \cite{Li2019GeneralizedZS}. Embedding-based methods, on the other hand, aim to learn a projection or embedding function to associate the low-level visual features of seen classes with their corresponding semantic representations and exploit the predicted semantic for classification. However, it is challenging to learn an explicit projection function between two spaces due to the distinctive properties of different modalities. What's more, data samples of seen and unseen classes are disjoint and their distributions are dissimilar, thus, learning a projection function using data samples from the seen classes without any adaptation to the unseen classes causes the domain shift problem \cite{Jia2020DeepUE, FuYanwei2015TransductiveMZ, Zhao2017ZeroshotLV}. In our work, we propose the Fine-grained Recognition branch and the Attribute Constraint branch to alleviate the above problems.


\noindent
\textbf{Semantic alignment in ZSL.}
The alignment between visual space and semantic space has always been a main issue of embedding-based ZSL methods, earlier works handled this issue from different aspects. For instance, \cite{Frome2013DeViSEAD, Akata2015EvaluationOO, RomeraParedes2015AnES, Zhang2015ZeroShotLV, Fu2018ZeroShotLO, Shigeto2015RidgeRH, Das2019ZeroshotIR, Zhang2019CoRepresentationNF} proposed to transform representations to the visual space or the semantic space for discriminating image samples. To mitigate the hubness problem \cite{Arora2018GeneralizedZL, Zhang2017LearningAD, Yang2018DissimilarityRL}, \cite{Zhang2017LearningAD, Yang2018DissimilarityRL, Paul2019SemanticallyAB, Zhang2020TowardsED} introduced a latent space for better alignment between visual and semantic space. Vyas et al. \cite{Vyas2020LeveragingSA} considered the relationships between different classes and used them as supplementary information of unseen samples. Recently, several methods \cite{Han2021ContrastiveEF, Wang2021TaskIndependentKM} got inspiration from self-supervised learning and leveraged the contrastive learning \cite{Tian2020ContrastiveMC, He2020MomentumCF} to obtain a better embedding. Specifically, Han et al. \cite{Han2021ContrastiveEF} learned a novel embedding space for instance discrimination via contrastive learning. Wang et al. \cite{Wang2021TaskIndependentKM} simultaneously learned task-specific and task-independent knowledge to produce transferable representations via contrastive learning. Although these methods learned better alignment to some extent, the core of the alignment between two spaces should not be the sample and its semantic description, but the image regions and semantic attributes. To achieve this region-attribute alignment, our work directly learn different attributes from different regions of image samples.

\noindent
\textbf{Part-based ZSL.}
Recently, several ZSL methods focused on leveraging detail information of visual features. Yu et al. \cite{Yu2018StackedSA} utilized attention mechanism to weigh different local image regions from class embeddings. Sylvain et al. \cite{sylvain2019locality} demonstrated the importance of locality and compositionality of image representations for zero-shot learning. Zhu et al. \cite{Zhu2019SemanticGuidedML} developed the attention method by learning multiple channel-wise part attentions. Huynh et al. \cite{Huynh2020FineGrainedGZ} used an attribute-based dense attention mechanism to alleviate the bias problem \cite{Zhang2019CoRepresentationNF}. To utilize the relationships between image regions, \cite{Xie2020RegionGE} exploited a region graph to learn better visual embedding, but they failed to directly map the attributes to the image regions. Xu et al. \cite{Xu2020AttributePN} learned a prototype for each attribute via utilizing the local features, they improved the locality of image representations but failed to directly exploit local features in ZSL inference. Similar to \cite{Xu2020AttributePN}, \cite{Liu2021GoalOrientedGE} exploited a gaze estimation method to predict the human gaze towards an image sample and transformed the gaze information to the attribute attention of different image regions, however, they still relied on global features in ZSL inference. In our work, we directly leverage the outputs of image encoder and then map the regions of visual features to corresponding semantic attributes. 

\section{The proposed method}
\subsection{Problem Setting}
The objective of ZSL is to classify images $\mathcal{X}$ into unseen classes \textit{U} by transferring the knowledge of seen classes \textit{S} through semantic description $\mathcal{A}$. In ZSL, the seen and unseen classes are disjoint, i.e. $\mathcal{Y}^s \cap \mathcal{Y}^u = \emptyset$.  We define the training set  $\mathcal{D}^{train}={\{ x^{s} \in \mathcal{X}^{s}, y^{s} \in \mathcal{Y}^{s}, a(y^{s}) \in \mathcal{A}^{s} \}}$ where $x^{s}$denotes a sample from seen classes $x^{s}$, $y^{s}$denotes the label of seen class which is available in training phase and $a(y^{s}) \in \mathbb{R}^K$ denotes the semantic description of a seen class which contains \textit{K} semantic attributes. In testing phase, we have access to the semantic description of unseen classes $\mathcal{A}^{u}$. ZSL aims to predict the label of image from unseen classes, i.e. $\mathcal{X}^{u} \rightarrow \mathcal{Y}^{u}$, while GZSL aims to predict the labels of images from both seen and unseen classes, i.e. $\mathcal{X} \rightarrow \mathcal{Y}^{s} \cup \mathcal{Y}^{u}$. In addition, we assume the access to GloVe \cite{Pennington2014GloveGV} representations of text descriptions of all the attributes. Specifically, for the $n$-th word in the $k$-th attribute, its word embedding is denoted as $e^k_n$.

\subsection{Overview}
As shown in Figure \ref{fig:framwork}, the Region Semantically Aligned Network (RSAN) consists of two subbranches: the Fine-grained Recognition branch (FR) and the Attribute Constraint branch (AC). Both branches share an image encoder which maps the image sample $\mathbf{x}$ to a feature map $v(\mathbf{x}) \in \mathbb{R}^{C \times H \times W}$, where $H$, $W$ and $C$ are the height, width and channel of the feature map. The FR branch is capable of mapping the image regions to the semantic attributes and using the predicted attributes to discriminate the samples from different classes. Moreover, the AC branch focuses on learning better image encoder via attributes regression with semantic knowledge.

\begin{figure}[t]
\centering

\includegraphics[scale=0.3]{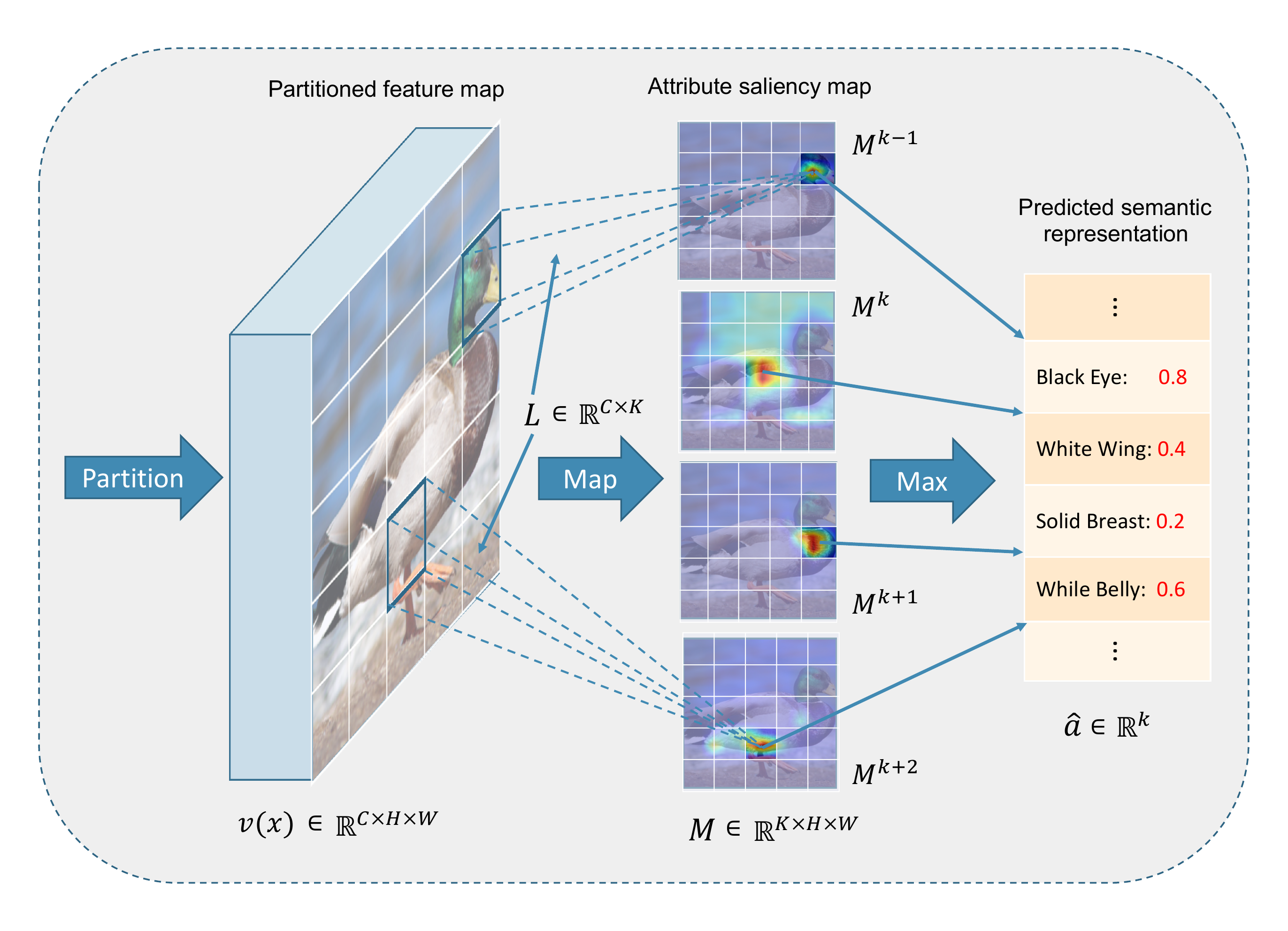}

\caption{Illustration of the region-based mapping process. The region-based mapping process generally consists of three operations among which “partition” and “map” together compute the attribute saliency over all the regions for each attribute through $P$. Then, the “max” operation selects regions of the highest attribute saliency to obtain the semantic representation of the sample.}
\label{fig:region_wise}
\end{figure}

\subsection{Fine-grained Recognition Branch}
To fully leverage the detail information during seen knowledge transfer, in FR branch, we first map the visual features of image regions to its semantic attributes and obtain the predicted semantic representation $\hat{a}$ (region-based mapping module), then we exploit a cosine space to classify the sample via its predicted semantic representation (cosine embedding module).  

\noindent
\textbf{Region-based mapping.}
A commonly used method in ZSL \cite{Frome2013DeViSEAD, Akata2015EvaluationOO, RomeraParedes2015AnES, Zhang2015ZeroShotLV, Fu2018ZeroShotLO} is to map the global visual feature of a sample to its semantic space and exploit the predicted semantic representation to classify the sample. Therefore, an appropriate mapping from the visual space to its semantic space is crucial for ZSL problems. Given the feature map $v(x) \in \mathbb{R}^{C \times H \times W}$ of an image $x$, previous methods \cite{Xu2020AttributePN, Liu2021GoalOrientedGE} firstly applied a global average pooling layer to get a global feature $g(x) \in \mathbb{R}^C$:
\begin{equation}
g(x)=\frac{1}{H \times W} \sum_{i=1}^{H} \sum_{j=1}^{W} v_{i, j}(x),
\label{eq:avgpool}
\end{equation}
where $v_{i, j}(x)$ is extracted from the feature $v(x)$ at spatial location $(i, j)$.

Then a projection matrix $V \in \mathbb{R}^{C \times K}$ is learned to project $g(x)$ to its semantic space $\hat{a}(x) \in \mathbb{R}^K$:
\begin{equation}
\hat{a}(x)=g(x)^{T} V.
\label{eq:gx}
\end{equation}

However, a specific attribute from the semantic description usually denotes part information of a sample, e,g., bill shape for a bird in CUB dataset \cite{Wah2011TheCB}, global features fail to accurately represent information for certain attribute since the attribute only corresponds to a region of the image and the redundant information from other regions plays a negative role under such a circumstance. What's more, each dataset contains several similar classes which can not be easily discriminated while using global features, e.g. chimpanzee and gorilla. When an unseen class shares a lot of semantic attributes with a seen class, the images of the unseen class would easily be classified into the similar seen class using the global features. Therefore, instead of global feature mapping, we propose a Region-based mapping module to directly map the visual features of image regions to its semantic attributes.   

As shown in Figure \ref{fig:region_wise}, given a visual feature map $v(\mathbf{x}) \in \mathbb{R}^{C \times H \times W}$, we naturally obtain $H \times W$ regions where each region $(i,j)$ is represented by a C-dimensional vector. Then we learn the significance level of each semantic attribute for each region. In detail, we exploit a fully-connected layer with parameter $P \in \mathbb{R}^{C \times K}$ to map visual feature of each region to the significance level of each semantic attribute. Finally we gather the attribute significance level for all regions together in spatial order and obtain an attribute saliency map $M \in \mathbb{R}^{K \times H \times W}$. 

It is obvious that fine-grain attributes should learn from the visual features of specific image regions, as for coarse-grain attributes, two reasons ensure the region-based mapping remains effective: 1. The visual features of image regions extracted by deep convolution network have a receptive field as large as the original image size theoretically. The region visual features actually contain the information from the whole image and focus mostly on the corresponding image regions. 2. It is unnecessary to consider every part of the coarse-grain attribute, focusing on one or a few discriminative regions is enough for knowledge transfer. To this end, we would like to constrain the unrelated region to have a low significance level for each attribute, especially for the regions that are far from the peak regions. Therefore, we exploit a concentrate loss\cite{Zheng2017LearningMC} on $M$ to regularize the attribute saliency map
\begin{equation}
\mathcal{L}_{Con}=\sum_{k=1}^{K} \sum_{i=1}^{H} \sum_{j=1}^{W} M_{i, j}^{k}\left[(i-\tilde{i_k})^{2}+(j-\tilde{j_k})^{2}\right],
\label{eq:lcon}
\end{equation}
where $(\tilde{i_k}, \tilde{j_k})$ denotes the location with the highest significance level in attribute saliency map for the k-th attribute. 

Given that the attribute saliency map shows the significance level of a region for a contain attribute, we can localize the attribute by finding the peak region of the attribute saliency map and the predicted value $\hat{a}_{k}$ for $k$-th attribute can be denoted as
\begin{equation}
\hat{a}_{k}=\max _{i, j} M_{i, j}^{k}.
\end{equation}

In section 4.3, we verify the effectiveness of our Region-based Mapping module via reporting a significant improvement on the baseline model in both ZSL and GZSL settings.

\noindent
\textbf{Cosine embedding.}
Semantic description is a useful information to discriminate samples from different classes. However, the image samples from certain class often fail to express all attribute information from their semantic descriptions due to various filming angles and the blocking of other objects. To tackle this problem, follow \cite{skorokhodov2021class,Liu2021GoalOrientedGE}, we propose a robust Cosine Embedding module to focus on the most discriminative attributes during classification. Specifically, we computer cosine similarity between the predicted semantic representation $\hat{a}$ and the ground truth semantic descriptions. Concretely, we define the classification loss for a given image $x$ with label $y$ as
\begin{equation}
\mathcal{L}_{Cls}=-\log \frac{\exp \left(cos\left(\hat{a}, a(y)\right) / \tau_{s}\right)}{\sum_{y^s \in \mathcal{Y}^{s}} \exp \left(cos\left(\hat{a}, a(y^s)\right) / \tau_{s}\right)},
\label{eq:ce}
\end{equation}
where $\tau_{s}>0$ is the temperature parameter and $a(y)$ denotes the ground truth semantic description of the label $y$ . In section 4.3, we validate that the cosine embedding helps to learn a robust discriminator for both ZSL and GZSL problem. 

\subsection{Attribute Constraint Branch}
Although the Fine-grained Recognition branch is able to localize semantic attributes for an input with the knowledge transferred from seen categories, the accuracy of localization cannot be guaranteed because other irrelevant visual features like "branch" of a tree may be mistaken as "leg" of a bird by our Region-based Mapping branch if we try to localize attribute about "leg" from a bird. This is because the image encoder is pretrained on other dataset which is not directly related to our task. Moreover, the problem of domain shift \cite{Jia2020DeepUE, FuYanwei2015TransductiveMZ, Zhao2017ZeroshotLV} further degrades the performance of the image encoder. Specifically, samples from two categories share the same attribute but the visual patterns of the attribute differ significantly. As a result, the Region-based Mapping module may fail to localize the attribute correctly. Therefore, we propose our Attribute Constraint branch (AC) to regularize the image encoder to extract robust and attribute-related visual features through attribute regressions with semantic knowledge.

\noindent
\textbf{Attribute-aware feature map.}
Feature maps obtained by the image encoder perform poorly for localization because the image encoder cannot sufficiently extract attribute-related visual patterns for ZSL problem and suffers severely from domain shift. We alleviate these two problems by obtaining a better visual representation of the image sample, i.e., attribute-aware feature map. Specifically, we add another convolution layer comprised of heuristic-initialized kernels after the image encoder and obtain estimated attributes of the sample. Then, we apply attribute regression between the estimated attributes and the ground truth attributes to enforce the feature map to encode semantic attributes. 
For the convolution layer, we formulate our attribute-kernels as follows:
\begin{equation}
    H=\left\{h_{k} \in \mathbb{R}^{C \times h \times w}\right\}_{k=1}^{K},
\end{equation}
where $h$, $w$ denotes the height and width of the attribute-kernel respectively, $C$ denotes the number of channels in $f(\mathbf{x})$, and $K$ denotes the number of attributes.

Interestingly, we do not initialize our attribute-kernel with random or normal distribution. Instead, we leverage semantic knowledge of each attribute as auxiliary information to initialize the attribute kernels, which aims to mitigate the problem of domain shift. As a result, the image encoder can tolerate visual variance of an attribute. Concretely, for the text description of the $k$-th attribute, we obtain its vector representation by averaging the word embedding of all the words appearing in the text:
\begin{equation}
    \mathbf{E}^k = \frac{1}{N^k}\sum_{n=1}^{N^k} \mathbf{e_n^k},
\end{equation}
where $N^k$ denotes the number of words in the text description of the $k$-th attribute, and $e_n^k \in \mathbf{R}^{d}$ denotes the GloVe \cite{Pennington2014GloveGV} representation of the $n$-th word in the former, which has a dimensionality of $d$.

Then, we initialize our attribute kernels with the aforementioned embedding by first applying a fully-connected layer to the embedding and then performing a "reshape" operation to adjust the parameters to the size of the kernels:
\begin{equation}
h_{k} = \operatorname{Reshape}({E}^k \mathbf{W}, h, w),
\label{eq:InitK}
\end{equation}
where $\operatorname{Reshape}$ denotes an operation to resize the parameters to $\mathbb{R}^{h \times w}$. Then the $k$-th attribute map is obtained by convolving the feature map with attribute-kernels:
\begin{equation}
    A^k = \sigma(\operatorname{Conv}(v(\mathbf{x}),h_{k})) \in \mathbb{R}^{C \times (H-h+1) \times (W-w+1)},
\end{equation}
where $\operatorname{Conv}(input,kernel)$ denotes a convolution operation with no padding and stride$=$1, $\sigma()$ denotes an activation function to provide non-linearity.

For the attribute regression, firstly we obtain the estimated $k$-th attribute value of $\mathbf{x}$ by applying a max pooling layer in the attribute map:
\begin{equation}
    a^{reg}_k =  \operatorname{Maxpool}(A^k).
\label{eq:attribute_proposal}
\end{equation}
Finally we utilize Mean Square Error (MSE) to minimize $L_{2}$-distance between $a^{reg}$ and its corresponding semantic description $a(\mathbf{y})$:
\begin{equation}
    \mathcal{L}_{Reg} = \|a^{reg}-a(y)\|_{2}^{2}.
    \label{eq:Lreg}
\end{equation}

\subsection{Joint Both Branches for ZSL Recognition}
As described above, the Region Semantically Aligned Network consists of two subbranches. Fine-grained Recognition branch is learned through optimizing the Classification loss $\mathcal{L}_{Cls}$ and the Concentrate loss $\mathcal{L}_{Con}$. Attribute Constraint branch is learned through optimizing the Regression loss $\mathcal{L}_{Reg}$. Thus, the overall objective can be written as follows:
\begin{equation}
\mathcal{L}=\mathcal{L}_{Cls}+\lambda_{1}\mathcal{L}_{Con} +\lambda_{2}\mathcal{L}_{Reg},
\end{equation}
where $\lambda_{1},\lambda_{2}$ are hyper-parameters. The joint of Fine-grain Recognition branch and Attribute Constraint  branch enables the model to correctly map the image regions of an unseen sample to its semantic attributes.

\noindent
\textbf{Zero-shot prediction.}
After the full model is trained, we exploit our trained Region-based Mapping module to predict the semantic representation of the image sample (denoted as $\psi_{RM}\left(\mathbf{x}^{u}\right)$) and then utilize the Cosine Embedding module to discriminate the test image samples. Given a test sample $\mathbf{x}$, in ZSL setting, the prediction $\hat{\mathbf{y}}$ is made by finding the best match ground truth semantic description in unseen classes via
\begin{equation}
\hat{\mathbf{y}}=\underset{\mathbf{y} \in \mathcal{Y}^{U}}{\arg \max } \cos \left( \psi_{RM}\left(\mathbf{x}\right), a(\mathbf{y})\right).
\end{equation}

Besides, GZSL settings use both seen and unseen samples for testing and suffer from the bias problem \cite{Zhang2019CoRepresentationNF, Chao2016AnES}. We exploit the calibrated stacking \cite{Chao2016AnES} to alleviate this problem. Therefore, the GZSL prediction can be defined as 
\begin{equation}
\hat{\mathbf{y}}=\underset{{y} \in \mathcal{Y}^{U} \cup \mathcal{Y}^{S}}{\arg \max }\left(\sigma \cos \left(\psi_{RM}\left(\mathbf{x}\right), a(\mathbf{y})\right)-\gamma \mathbb{I}\left[{\mathbf{y}} \in \mathcal{Y}^{S}\right]\right),
\end{equation}
where $\gamma$ is a calibration factor and $\mathbb{I}=1$ while ${y}$ is a seen class and 0  while ${y}$ is an unseen class. To avoid vanishing gradient problem, we also adopt a scaling factor $\sigma$ during the cosine similarity computation.

\section{Experiment}
We evaluate our proposed method on three benchmark datasets, including Caltech-UCSD Birds 200 (CUB) \cite{Wah2011TheCB}, Animals with Attributes 2 (AWA2) \cite{XLSA18} and Scene UNderstanding (SUN) database \cite{Patterson2012SUNAD}. Below, we discuss the datasets, evaluation metrics and implementing details. After that, we compare our method with state-of-the-art ZSL methods. Then we perform ablation studies to demonstrate the effectiveness of different components in our model. Next we analyse the effect of hyper-parameters on the performance of our method. Finally we examine our method's abilities to discriminate between similar categories and to localize semantic attributes accurately.
\subsection{Experiment Setup}
\textbf{Datasets.}
Following \cite{Liu2021GoalOrientedGE, Xu2020AttributePN}, we conduct experiments on three benchmark datasets: CUB, AWA2, SUN. 
CUB \cite{Wah2011TheCB} contains fine-grained bird images from 150 seen and 50 unseen classes with 312 attributes. SUN \cite{Patterson2012SUNAD} is a dataset of visual scenes having 645 seen and 72 unseen classes with 102 attributes and it has the largest number of classes among all datasets. However, it only contains 16 training images per class due to its small overall training set. AWA2 \cite{XLSA18} has been proposed for animal classification with 40 seen and 10 unseen classes and each of which is described by 85 attributes. It has a medium size of 37,322 samples in total. For CUB, SUN, AWA2, we follow the proposed training, validation and testing splits in \cite{XLSA18}.


\noindent
\textbf{Evaluation protocols.}
The performance of ZSL is evaluated by average perclass Top-1 (\textbf{T1}) accuracy. In GZSL, since the test set is composed of seen and unseen images, the Top-1 accuracy evaluated respectively on seen classes, denoted as \textbf{S}, and unseen classes, denoted as \textbf{U}. Their harmonic mean \cite{XLSA18}, defined as $\mathbf{H} = (2 \times \mathbf{S} \times \mathbf{U})/(\mathbf{S} + \mathbf{U})$.

\noindent
\textbf{Implementing details.} 
Following the canonical setting in \cite{Liu2021GoalOrientedGE}, We use a pretrained ResNet-101 with the input size of $448 \times 448$ for feature extraction. We extract a feature map at the last convolutional layer whose size is $14 \times 14 \times 2048$ and treat it as a set of features from $14 \times 14$ regions. The SGD optimizer is adopted in the model training. The momentum is set to $0.9$, and the weight decay is $10^{-5}$. The learning rate is initialized as $10^{-3}$ and decreased every ten epochs by a factor of $0.5$. Other hyper parameters in our model are obtained by grid search on the validation set \cite{XLSA18}. We set $\lambda_{1}$ and $\lambda_{2}$ as $0.1$, $1.0$ for three datasets. For attribute kernel sizes, we choose 1, 5, 3 for CUB, SUN and AWA2 respectively. The factor for Calibrated Stacking is set to 0.7 for CUB and SUN, and 3.75 for AwA2. We use an episode-based training method to sample $M$ categories and $N$ images for each category in a minibatch, we iterate 300 batches for each epoch, and train the model 20 epochs. We set $M = 16$ and $N = 2$ for all three datasets.

\begin{center}
\begin{table*}[!htbp]
\caption{Results (\%) of the state-of-the-art ZSL and GZSL. The first part is generative methods, the second part is semantic embedding methods. The best results are marked in \textcolor{red}{red}.}
\label{table:SOTA} 
\setlength{\tabcolsep}{8.5pt}
\begin{tabular}{|c|c|c|c|c|c|c|c|c|c|c|c|c|}
\hline
\multirow{3}{*}{Method} & \multicolumn{4}{c|}{CUB}         & \multicolumn{4}{c|}{AWA2}        & \multicolumn{4}{c|}{SUN}         \\ \cline{2-13} 
                        & ZSL  & \multicolumn{3}{c|}{GZSL} & ZSL  & \multicolumn{3}{c|}{GZSL} & ZSL  & \multicolumn{3}{c|}{GZSL} \\ \cline{2-13} 
                        & T1   & S       & U      & H      & T1   & S       & U      & H      & T1   & S       & U      & H      \\ \hline
f-CLSWGAN(CVPR'18) \cite{Xian2018FeatureGN}      & 57.3 & 57.7    & 43.7   & 49,7   & 68.2 & 61.4    & 57.9   & 59.6   & 60.8 & 36.6    & 42.6   & 39.4   \\
LisGAN(CVPR'19) \cite{Li2019LeveragingTI}         & 58.8 & 57.9    & 46.5   & 51.6   & 70.6 & 76.3    & 52.6   & 62.3   & 61.7 & 37.8    & 42.9   & 40.2   \\
OCD-CVAE(CVPR'20) \cite{Keshari2020GeneralizedZL}      & 60.3 & 59.9    & 44.8   & 51.3   & 71.3 & 73.4    & 59.5   & 65.7   & 63.5 & \textcolor{red}{42.9}     & 44.8   & 43.8   \\
LsrGAN(ECCV'20)\cite{Xie2020RegionGE}         & 60.3 & 59.1    & 48.1   & 53.0   & -    & -       & -      & -      & 62.5 & 37.7    & 44.8   & 40.9   \\
Composer(NeurIPS'20) \cite{Huynh2020CompositionalZL}    & 69.4 & 56.4    & 63.8   & 59.9   & 71.5 & 77.3    & 62.1   & 68.8   & 62.6 & 22.0    & \textcolor{red}{55.1}   & 31.4   \\
CE-GZSL(CVPR'21) \cite{Han2021ContrastiveEF}        & 77.5 & 66.8    & 63.9   & 65.3   & 70.4 & 78.6    & 63.1   & 70.0   & 63.3 & 38.6    & 48.8   & \textcolor{red}{43.1}   \\
GCM-CF(CVPR'21) \cite{Yue2021CounterfactualZA}         & -    & 59.7    & 61.0   & 60.3   & -    & 75.1    & 60.4   & 67.0   & -    & 37.8    & 47.9   & 42.2   \\\hline\hline
ALE(TPAMI'16) \cite{Akata2016LabelEmbeddingFI}          & 54.9 & 62.8    & 23.7   & 34.4   & 59.9 & 76.1    & 16.8   & 27.5   & 58.1 & 33.1    & 21.8   & 26.3   \\ 
AREN(CVPR'19) \cite{Xie2019AttentiveRE}           & 71.8 & 69.0    & 63.2   & 66.0   & 67.9 & 79.1    & 54.7   & 64.7   & 60.6 & 32.3    & 40.3   & 35.9   \\
DAZLE(CVPR'20) \cite{Huynh2020FineGrainedGZ}          & 65.9 & 59.6    & 56.7   & 58.1   & -    & 75.7    & 60.3   & 67.1   & -    & 24.3    & 52.3   & 33.2   \\
DVBE(CVPR'20) \cite{Min2020DomainAwareVB}           & -    & 73.2    & 64.4   & 68.5   & -    & 77.5    & 62.7   & 69.4   & -    & 41.6    & 44.1   & 42.8   \\
RGEN(ECCV'20) \cite{Xie2020RegionGE}           & 76.1 & 73.5    & 60.0   & 66.1   & \textcolor{red}{73.6} & 76.5    & \textcolor{red}{67.1}  & 71.5   & 63.8 & 31.7    & 44.0   & 36.8   \\
APN(NeurIPS'20) \cite{Xu2020AttributePN}         & 72.0 & 69.3    & 65.3   & 67.2   & 68.4 & 78.0    & 56.5   & 65.5   & 61.6 & 34.0    & 41.9   & 37.6   \\
Class-Norm(ICLR'21) \cite{skorokhodov2021class}     & -    & 50.7    & 49.9   & 50.3   & -    & 73.4    & 63.1   & 67.8   & -    & 41.6    & 44.7   & \textcolor{red}{43.1}   \\
GEM-ZSL(CVPR'21) \cite{Liu2021GoalOrientedGE}        & 77.8 & 77.1    & 64.8   & 70.4   & 67.3 & 77.5    & 64.8   & 70.6   & 62.8 & 35.7    & 38.1   & 36.9   \\\hline
RSAN(ours)              & \textcolor{red}{79.7} & \textcolor{red}{78.5}    & \textcolor{red}{67.6}   & \textcolor{red}{72.6} & 69.9 & \textcolor{red}{80.4}  & 65.0   & \textcolor{red}{71.8}   & \textcolor{red}{64.9} & 34.0    & 43.1   & 38.0  \\ \hline
\end{tabular}
\end{table*}
\end{center}

\subsection{Comparison with the State-of-the-Art Models}
We compare RSAN with two groups of state-of-the-art models. On the one hand, f-CLSWGAN \cite{Xian2018FeatureGN}, LisGAN \cite{Li2019LeveragingTI}, OCD-CVAE \cite{Keshari2020GeneralizedZL}, LsrGAN \cite{Xie2020RegionGE}, Composer \cite{Huynh2020CompositionalZL}, CE-GZSL \cite{Han2021ContrastiveEF} and GCM-CF \cite{Yue2021CounterfactualZA} learn generative models to approximate the distribution of class images as a function of class semantic descriptions. Thus, given semantic descriptions of unseen classes, these models augments features of seen classes with generated features from the unseen ones and learn a discriminative classifier in the fully supervised setting. On the other hand, ALE \cite{Akata2016LabelEmbeddingFI}, AREN \cite{Xie2019AttentiveRE}, DAZLE \cite{Huynh2020FineGrainedGZ}, DVBE \cite{Min2020DomainAwareVB}, RGEN \cite{Xie2020RegionGE}, APN \cite{Xu2020AttributePN}, DCEN \cite{Wang2021TaskIndependentKM}, Class-Norm \cite{skorokhodov2021class} and GEM-ZSL \cite{Liu2021GoalOrientedGE} through various methods, embed the visual features of the test samples into a semantic representation aligned with the attribute space where the final classification is performed.

Table \ref{table:SOTA} shows the results of our RSAN and the methods mentioned above on three datasets. Our RSAN achieves the state-of-the-art or competitive results in both ZSL and GZSL settings. On CUB dataset, RSAN outperforms all the compared methods with a large margin in both ZSL and GZSL settings. Since CUB is a more challenging fine-grained dataset which requires local discriminative attributes, the results prove the effectiveness of our model. For AWA2 dataset, our RSAN can achieve state-of-the-art result in GZSL settings. We also report competitive results on ZSL settings. On SUN dataset, the feature generation based models significantly outperform most other methods in GZSL settings. As SUN dataset contains more than 700 categories, the generative model can bring more features for generalization to unseen classes. However, generative-based methods require ground truth unseen semantic description during training which is unrealistic for real world applications. Compared with the other non-generation based methods in GZSL settings, the performance of our RSAN is competitive. Moreover, thanks to the significant effect of our Region-based Mapping, RSAN still outperforms other methods in ZSL settings.

\begin{center}
\begin{table}[!htbp]
\caption{Results (\%) of ZSL and GZSL ablation study on CUB, SUN and AWA2. The baseline is the image encoder with global average pooling layer followed by a linear layer and a dot product to compute cross-entropy loss. RM and CE denote Region-based Mapping module without the concentrate loss and Cosine Embedding module respectively.}
\label{table:abDown}
\begin{tabular}{|c|c|c|c|c|c|c|}
\hline \multirow{2}{*} { Components } & \multicolumn{2}{|c|} { CUB } & \multicolumn{2}{|c|} { SUN } & \multicolumn{2}{|c|} { AWA2 } \\
\cline { 2 - 7 } & T1 & H & T1 & H & T1 & H \\
\hline Baseline & $58.8$ & $51.3$ & $56.1$ & $30.8$ & $51.5$ & $57.7$ \\
$+$RM (without $\mathcal{L}_{Con}$) & $\mathbf{71.7}$ & $\mathbf{65.6}$ & $\mathbf{60.3}$ & $\mathbf{28.0}$ & $\mathbf{63.1}$ & $\mathbf{65.2}$ \\
$+\mathcal{L}_{Con}$ & $74.6$ & $66.8$ & $61.4$ & $30.9$ & $65.5$ & $69.6$ \\
$+$CE & $76.3$ & $70.8$ & $62.8$ & $30.5$ & $67.4$ & $70.5$ \\
\hline
\end{tabular}
\end{table}
\end{center}

\begin{center}
\begin{table}[!htbp]
\caption{Results (\%) of ZSL and GZSL ablation study on CUB, SUN and AWA2. Upon the architecture of an image encoder and the Fine-grained Recognition branch, which is the baseline of this experiment, we evaluate the effectiveness of the components of Attribute Constraint branch. 
}
\label{table:abUp}
\setlength{\tabcolsep}{3.0pt}
\begin{tabular}{|c|c|c|c|c|c|c|}
\hline \multirow{2}{*} { Components } & \multicolumn{2}{|c|} { CUB } & \multicolumn{2}{|c|} { SUN } & \multicolumn{2}{|c|} { AWA2 } \\
\cline { 2 - 7 } & T1 & H & T1 & H & T1 & H \\
\hline Baseline & $76.3$ & $70.8$ & $62.8$ & $30.5$ & $67.4$ & $70.5$ \\
$+\mathcal{L}_{Reg}$ & $77.6$ & $71.8$ & $63.7$ & $31.0$ & $68.5$ & $71.0$ \\ \hline
$+$Semantic knowledge (RSAN) & $\mathbf{79.7}$ & $\mathbf{72.6}$ & $\mathbf{64.9}$ & $\mathbf{38.0}$ & $\mathbf{69.9}$ & $\mathbf{71.8}$ \\
\hline
\end{tabular}
\end{table}
\end{center}

\subsection{Ablation Study}
\textbf{Fine-grained Recognition branch.} 
In Table \ref{table:abDown}, we illustrate the effect of our Fine-grained Recognition branch. First, we respectively evaluate two kinds of mapping methods, global feature mapping, i.e., applying global average pooling to the feature map followed by a linear layer defined in Eq. (\ref{eq:avgpool}) and Eq. (\ref{eq:gx}) and Region-based Mapping as stated in section 3.3. Then, we complete our Region-based Mapping module with $\mathcal{L}_{Con}$, a constraint that provides locality for each attribute to demonstrate its effect. Finally, we compare Cosine Embedding module defined in Eq. (\ref{eq:ce}) that tolerates the intra-class variance with its counterpart dot product to show how it improves the performance. To obtain a fair and general evaluation of every module, average perclass Top-1 (\textbf{T1}) accuracy for ZSL and harmonic mean for GZSL are adopted.

The results show that our method is highly effective: Region-based Mapping module (without concentrate loss) has improved (\textbf{T1}) accuracy for ZSL by $9.6\%$ among which CUB  contributes the most. This agrees with our expectation and the nature of the datasets: CUB is a fine-grain dataset with easily localized attributes like "bill\_shape:needle", thus it perfectly suits for our Region-based Mapping; AWA2 shares similar type of attributes as CUB but it also has holistic attributes which could be hard to localize and detect, e.g, "domestic" and "solitary"; most of the attributes in SUN are abstract and descriptive texts like "socializing" and "playing", but our method still improves the (\textbf{T1}) accuracy by $4\%$. Then we evaluate the effectiveness of $\mathcal{L}_{Con}$, results on AWA2 show that it has improved the (\textbf{T1}) accuracy by $2.4\%$ and harmonic mean by $4.4\%$. At last, we verify the effectiveness of Cosine Embedding module. Results show that it has improved the (\textbf{T1}) accuracy by $1.7\%$ and harmonic mean by $4.0\%$ on CUB.

\noindent
\textbf{Attribute Constraint branch.}
In Table \ref{table:abUp}, we demonstrate the effect of our Attribute Constraint branch. We use the architecture mentioned above, i.e., an image encoder with a complete Fine-grained Recognition branch as baseline. Upon that, we examine the effectiveness of the components in Attribute Constraint branch. At first, we just initialize our attribute kernels with no heuristic semantic knowledge, only a regression loss $\mathcal{L}_{Reg}$ in Eq. (\ref{eq:Lreg}) is applied to guide the learning of image encoder. Then we complete our model with a language-prior-enabled initialization as described in Eq. (\ref{eq:InitK}). We use the same evaluation protocol as the former ablation study.
\begin{figure}[ht]
\centering
\subfigure[\textbf{T1} accuracy]{
\begin{minipage}[t]{0.5\linewidth}
\centering
\includegraphics[scale=0.25]{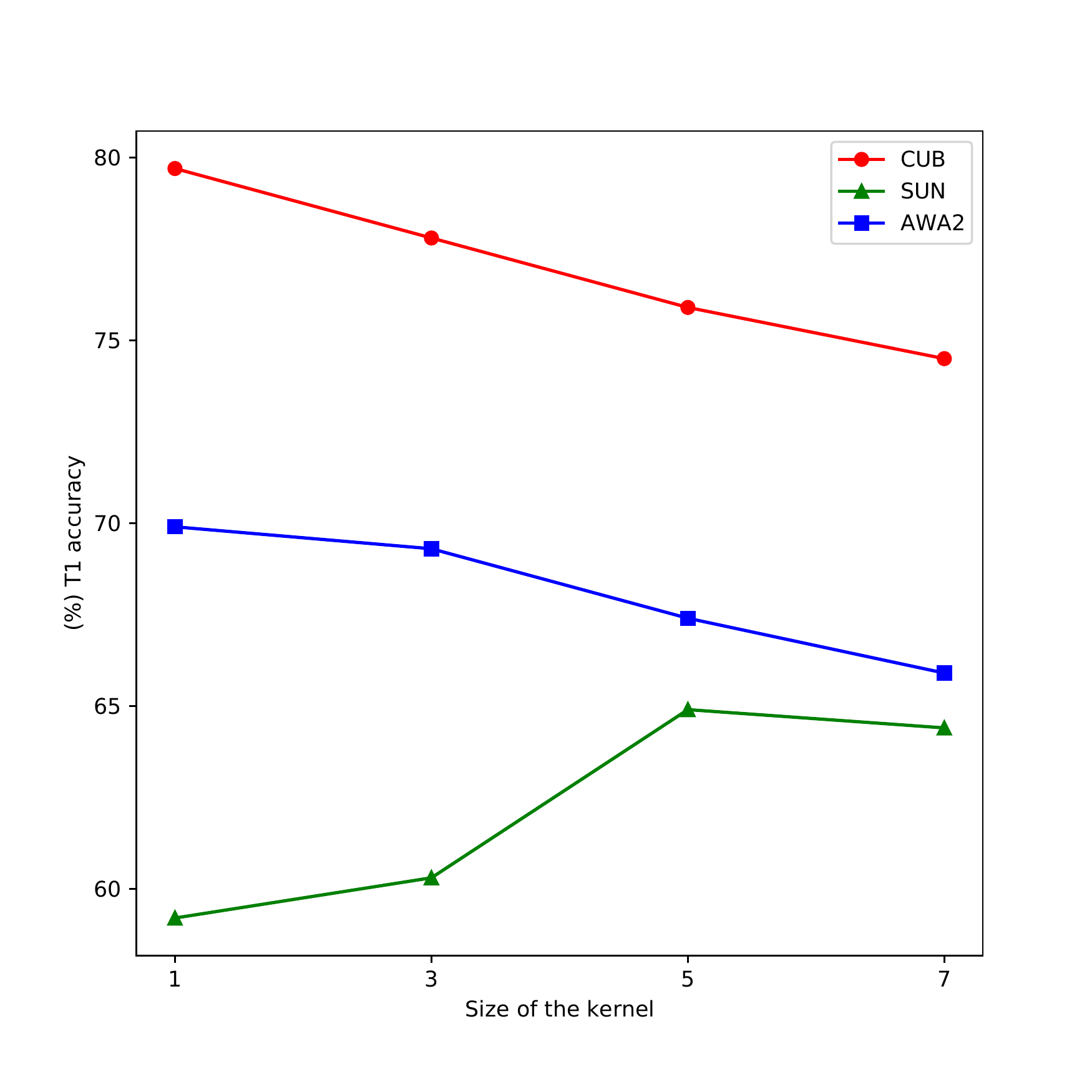}
\label{figure:kernel_t1}
\end{minipage}
}%
\subfigure[\textbf{H} harmonic mean]{
\begin{minipage}[t]{0.5\linewidth}
\centering
\includegraphics[scale=0.25]{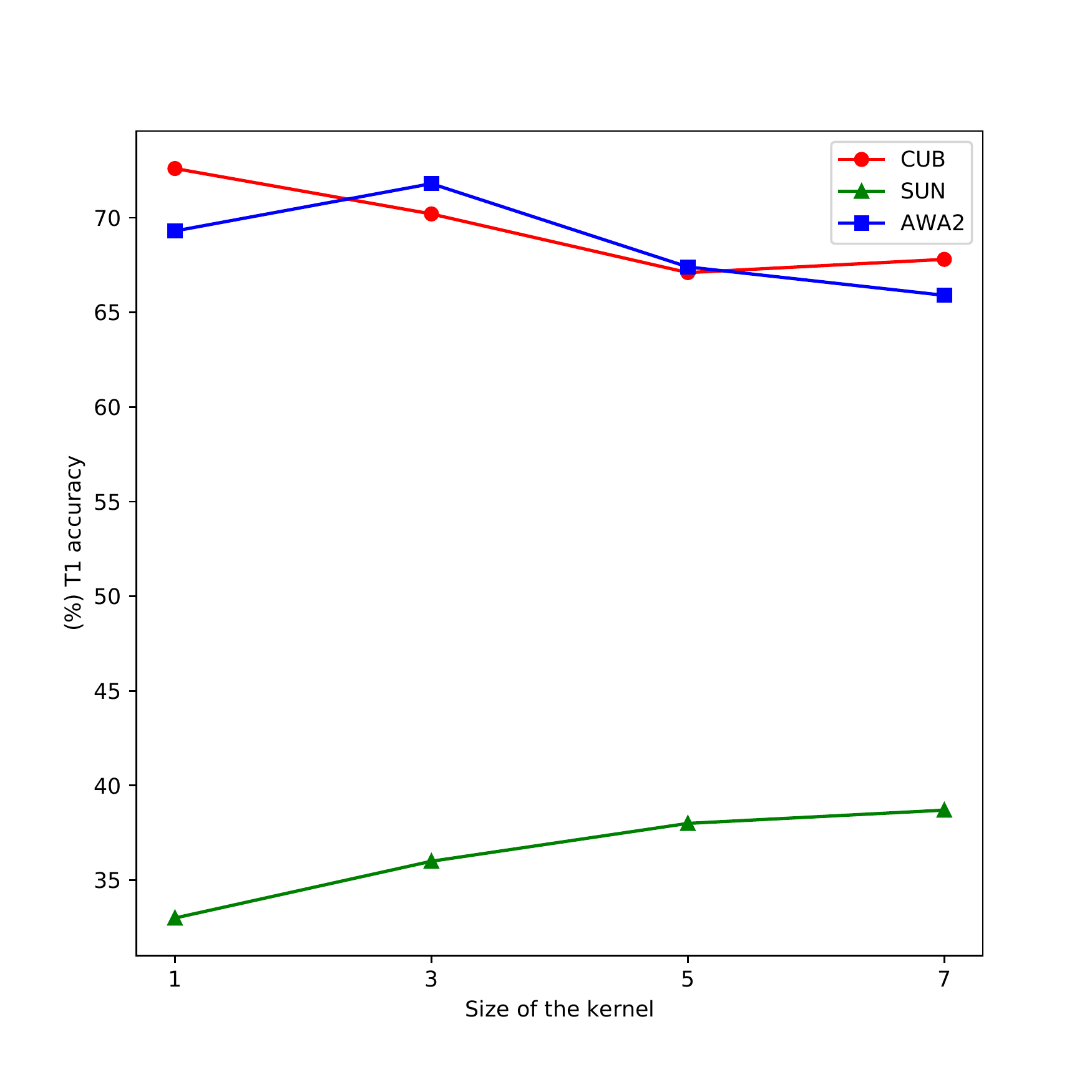}
\label{figure:kernel_H}
\end{minipage}
}%
\centering
\caption{the performance of RSAN on three datasets under the influence of different sizes of kernels 1, 3, 5 and 7. Figure \ref{figure:kernel_t1} is the result evaluated by T1 accuracy while Figure \ref{figure:kernel_H} is the result evaluated by (H) harmonic mean}
\label{fig:kernle_size}
\end{figure}

Through experimental results, our Attribute Constraint branch has been proved to be effective. Since $\mathcal{L}_{Reg}$ enables the image encoder to extract attribute-related visual features, the Fine-grained Recognition branch could handle the feature map more easily. Therefore, there is steady and considerable improvement among three datasets. Finally, when semantic knowledge is added to this branch, performance among three datasets has been significantly improved. Specifically, the harmonic mean on SUN has been improved by $7.0\%$. This agrees with the nature of the dataset given that SUN is a dataset that suffers the most from domain shift problem.

\subsection{Parameter Analysis}
\textbf{Effect of the size of attribute kernels.}
In section 3.4, we learn a set of attribute kernels to guide the learning of image encoder. Now we evaluate the effect of attribute kernel size on the performance of our model. As stated in section 4.4, we expect that because CUB has localized attribute, a relatively smaller kernel size is appropriate, and for AWA2 and SUN, an increasingly larger kernel size should be applied. Figure \ref{fig:kernle_size} shows under different size of kernels the  (\textbf{T1}) accuracy and harmonic mean of our model. It confirms our assumption. For the CUB, AWA2 and SUN datasets, they achieve their best performance in (\textbf{T1}) accuracy with kernel size 1, 3 and 5, respectively, and their best harmonic mean with kernel size 1, 3 and 7.

\noindent
\textbf{Training method analysis.}
An episode-based training method is used in our experiments to make the model gain better generalization ability. For each mini-batch, we sample $M$ categories and $N$ images for each category. We vary the value of $M$ with $\{8, 12, 16\}$ and the value of N with $\{2, 3, 4\}$, and observe the  (\textbf{T1}) accuracy under these values. To further analyze the performance of the episode-based training method, we compare its performance with the random sampling training method with a mini-batch of 64. Table \ref{table:episode} shows that the episode-based training method has better performance than the random sampling training method. The model can be generalized to the recognition of all categories (seen and unseen categories) only by learning the seen categories. When M = 16 and N = 2, the model can get the highest accuracy.

\begin{center}
\begin{table}
\caption{Influence of training method on ZSL results (\%). $\mathcal{R}$ represents random sampling training method with mini-batch of $64$, $\mathcal{E}$ represents episode-based training method.}
\label{table:episode}
\begin{tabular}{c|cc|ccc}
\hline Training Method & $M$ -way & $N$ -shot & CUB & SUN & AWA2 \\
\hline \hline $\mathcal{R}$ & $-$ & $-$ & $72.0$ & $59.2$ & $63.5$ \\
\hline & 8 & 2 & $74.4$ & $56.7$ & $67.7$ \\
& 8 & 3 & $74.0$ & $53.1$ & $66.8$ \\
& 8 & 4 & $74.5$ & $51.4$ & $66.4$ \\
$\mathcal{E}$ & 12 & 2 & $73.4$ & $61.1$ & $67.2$ \\
& 12 & 3 & $76.2$ & $63.6$ & $66.9$ \\
& 12 & 4 & $75.1$ & $63.9$ & $67.1$ \\
& 16 & 2 & $\textcolor{red}{79.7}$ & $\textcolor{red}{64.9}$ & $\textcolor{red}{69.9}$ \\
& 16 & 3 & $72.4$ & $63.9$ & $66.9$ \\
& 16 & 4 & $72.3$ & $59.4$ & $68.2$ \\
\hline
\end{tabular}
\end{table}
\end{center}

\begin{figure}
\centering
\subfigure[bobcat]{
\begin{minipage}[t]{0.3\linewidth}
\centering
\includegraphics[scale=0.35]{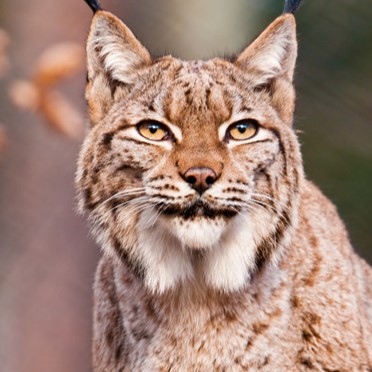}
\label{figure:bobcat}
\end{minipage}
}%
\subfigure[leopard]{
\begin{minipage}[t]{0.3\linewidth}
\centering
\includegraphics[scale=0.35]{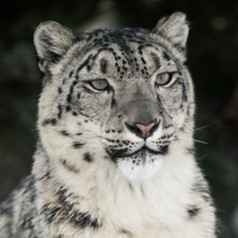}
\label{figure:leopard}
\end{minipage}
}%
\subfigure[lion]{
\begin{minipage}[t]{0.3\linewidth}
\centering
\includegraphics[scale=0.35]{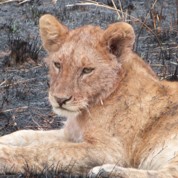}
\label{figure:lion}
\end{minipage}
}%
\centering
\caption{Three categories of "cats" from AWA2 share a lot visual features and can be hard to distinguish. "Bobcat" is from the unseen domain while others are all from the seen domain.}
\label{fig:similar}
\end{figure}

\begin{figure}
    \centering
    \includegraphics[width=0.4\textwidth]{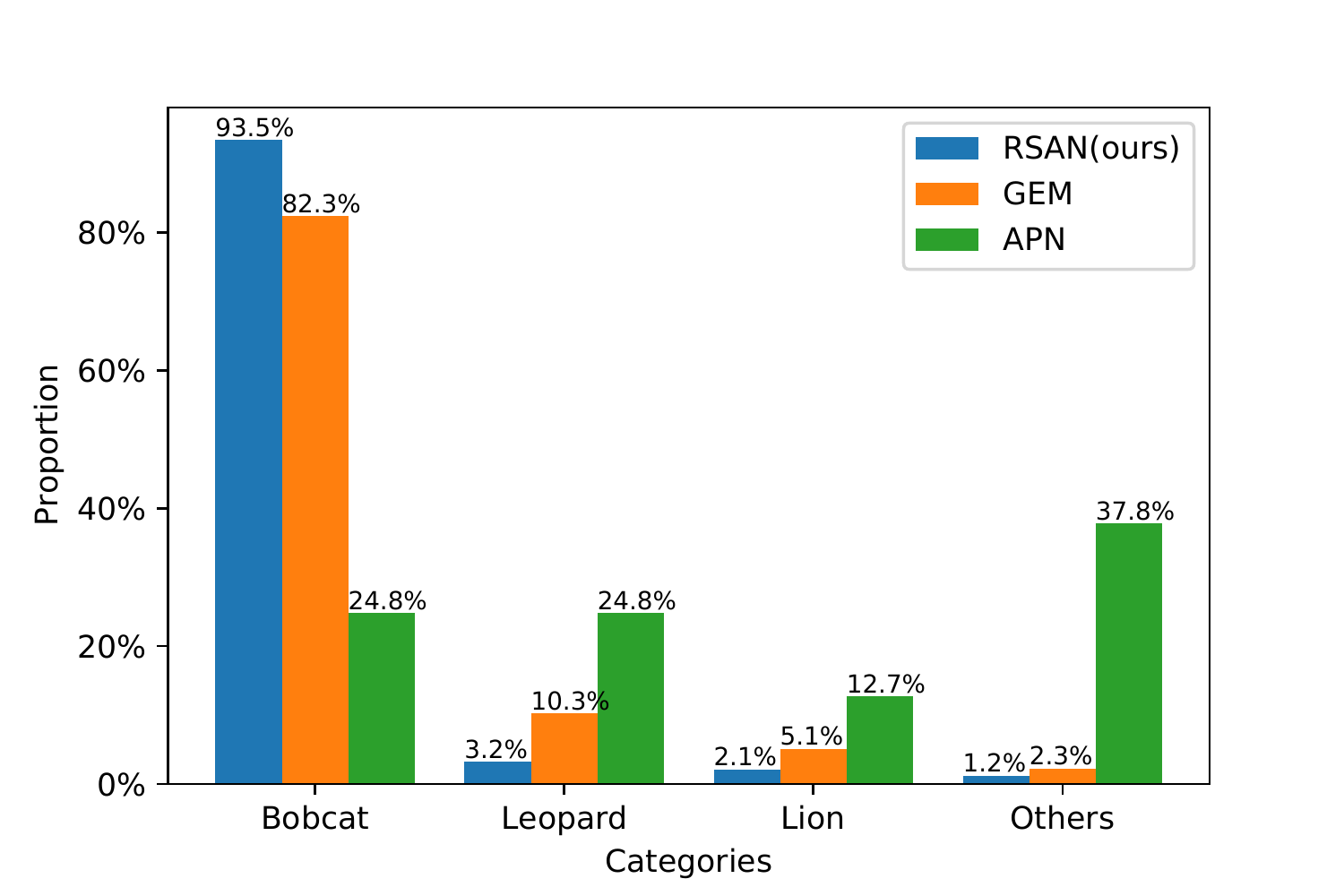}
    \caption{(\%) proportion that samples of the unseen category "bobcat" are classified into its own category and three similar categories from seen domain. "Other" denotes the rest of AWA2 categories, i.e., dissimilar categories.}
    \label{fig:compare}
\end{figure}

\begin{figure*}[htbp]
\centering

\subfigure[Original input]{
\begin{minipage}[t]{0.15\linewidth}
\centering
\includegraphics[width=1\linewidth, height=1\linewidth]{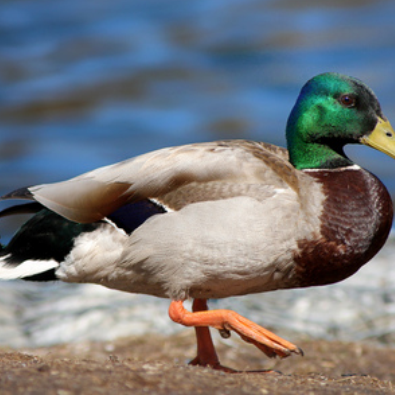}\vspace{1mm}
\includegraphics[width=1\linewidth, height=1\linewidth]{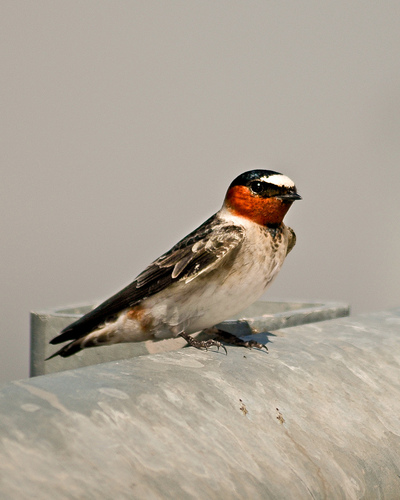}\vspace{1mm}
\includegraphics[width=1\linewidth, height=1\linewidth]{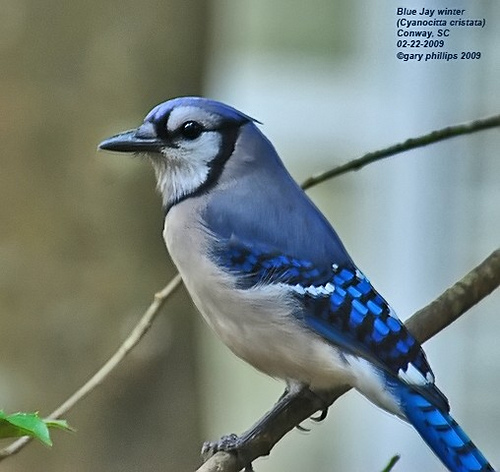}
\label{figure:ori}
\end{minipage}
}%
\subfigure[eye]{
\begin{minipage}[t]{0.15\linewidth}
\centering
\includegraphics[width=1\linewidth]{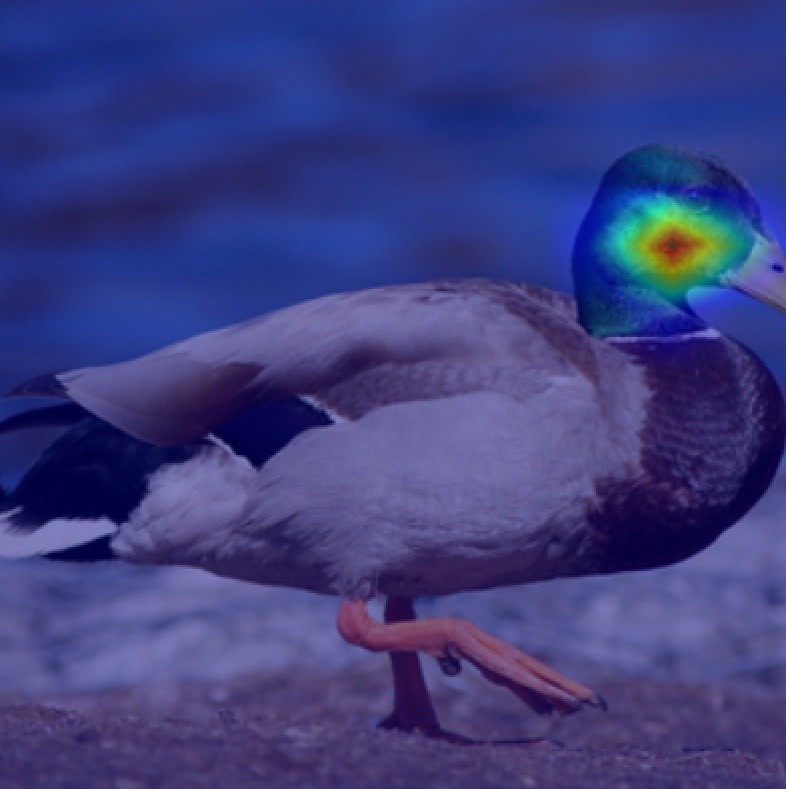}\vspace{1mm}
\includegraphics[width=1\linewidth]{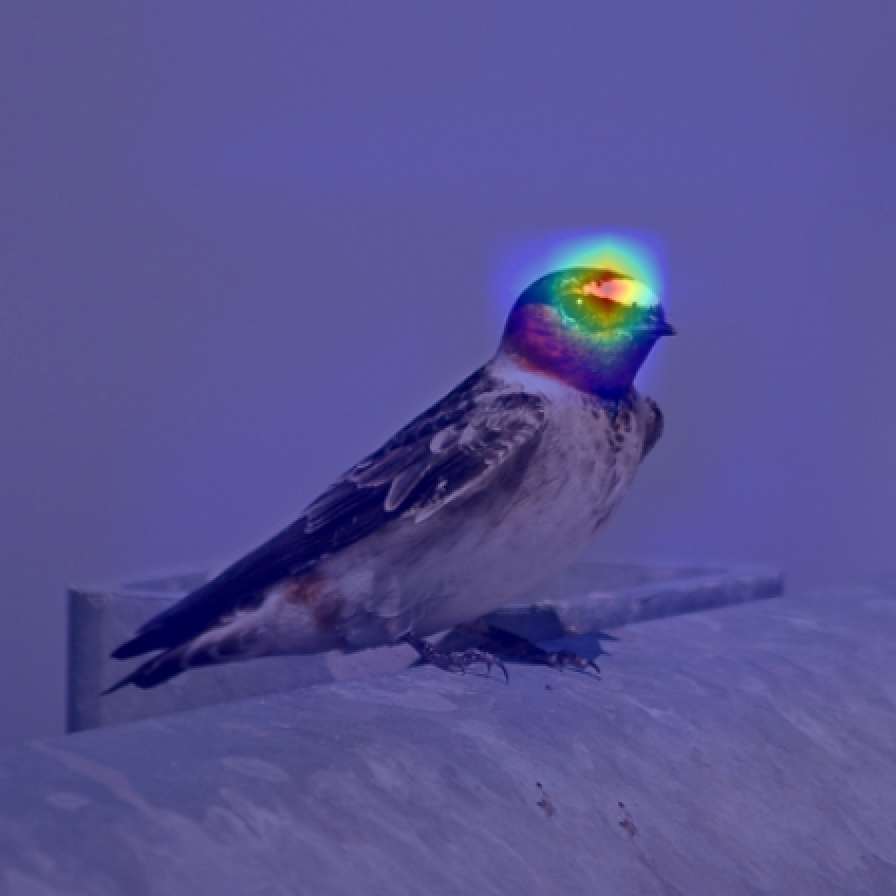}\vspace{1mm}
\includegraphics[width=1\linewidth]{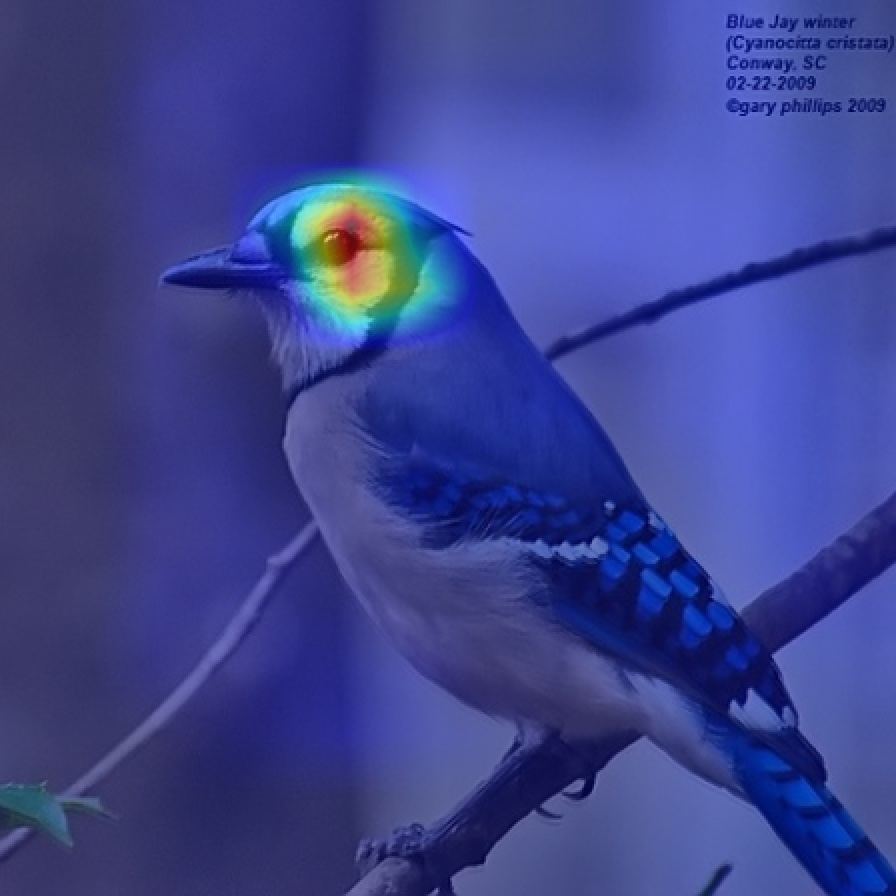}
\label{figure:eye}
\end{minipage}
}%
\subfigure[nape]{
\begin{minipage}[t]{0.15\linewidth}
\centering
\includegraphics[width=1\linewidth]{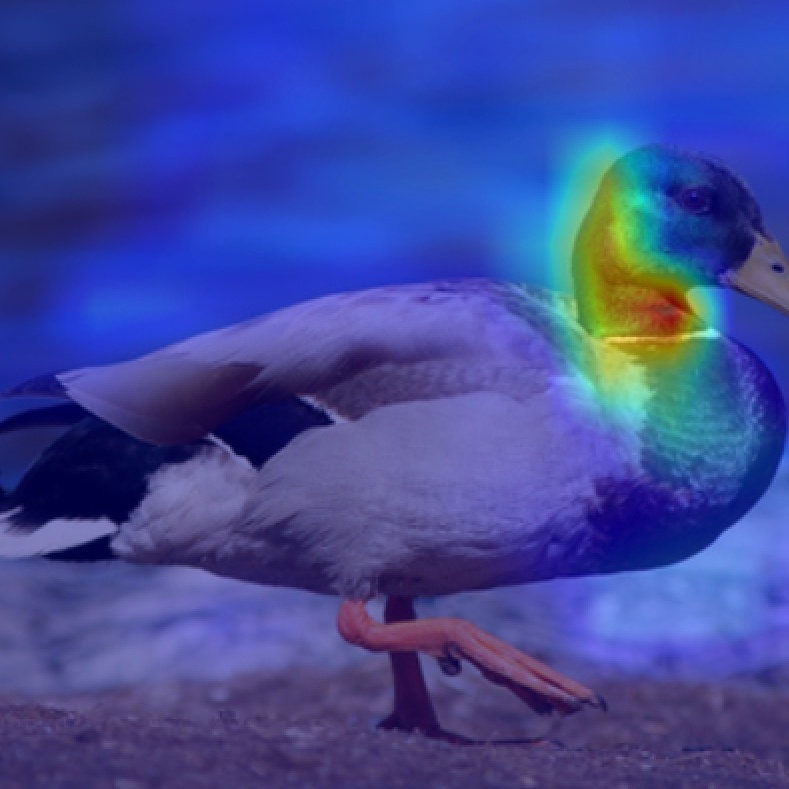}\vspace{1mm}
\includegraphics[width=1\linewidth]{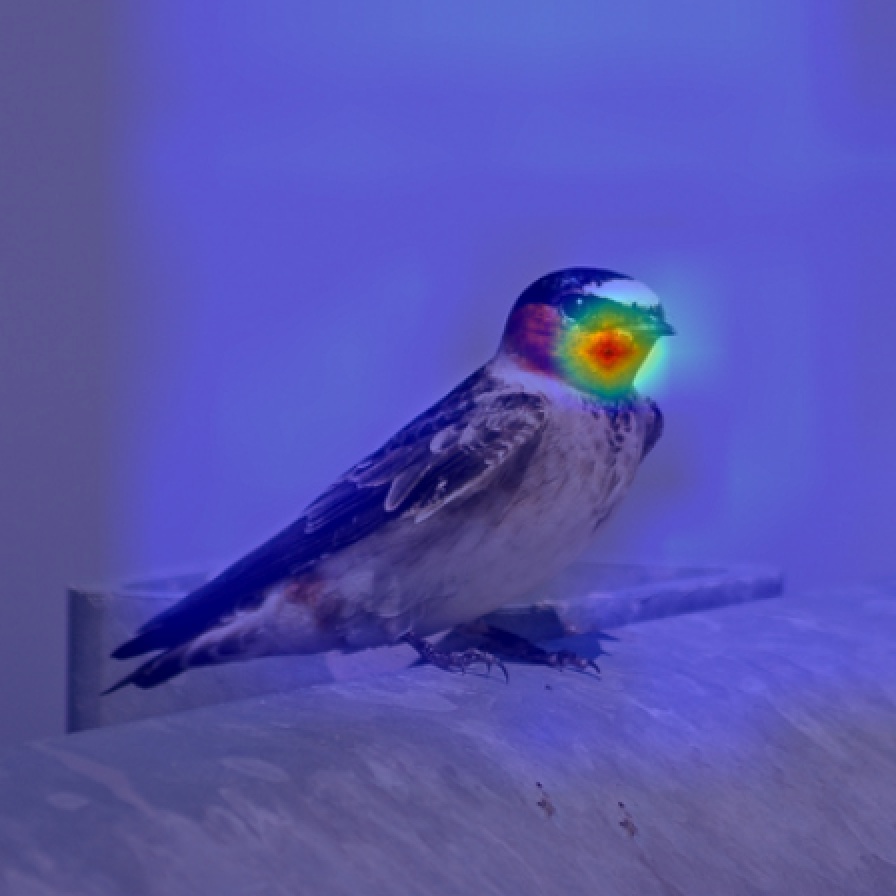}\vspace{1mm}
\includegraphics[width=1\linewidth]{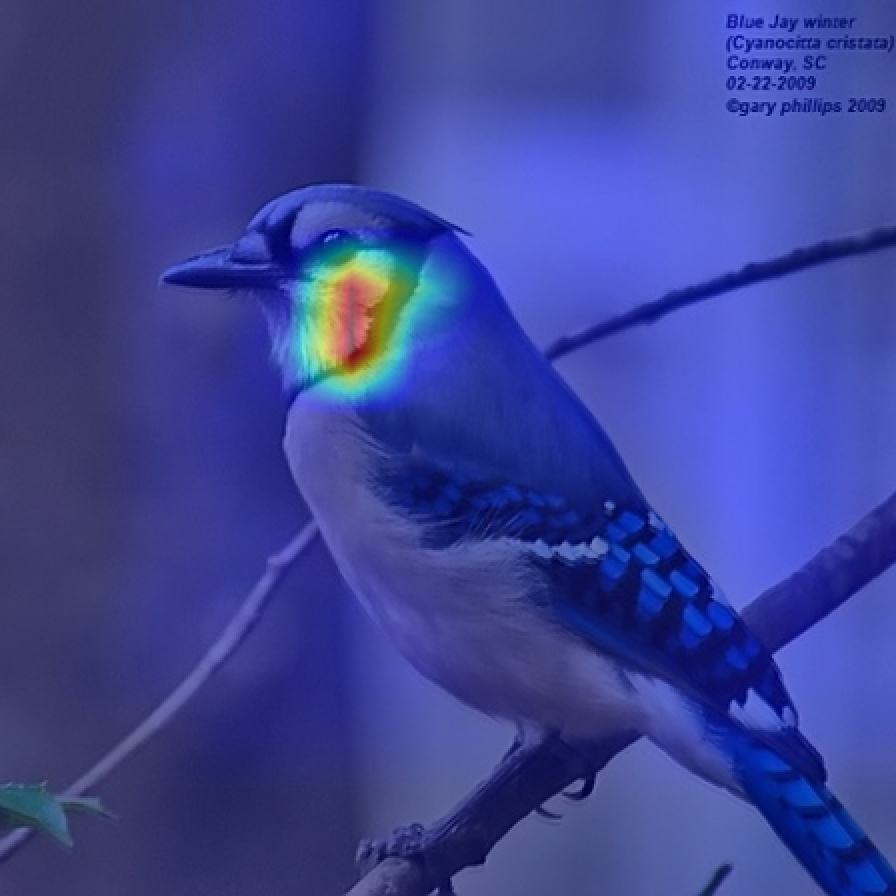}
\label{figure:nape}
\end{minipage}
}%
\subfigure[breast]{
\begin{minipage}[t]{0.15\linewidth}
\centering
\includegraphics[width=1\linewidth]{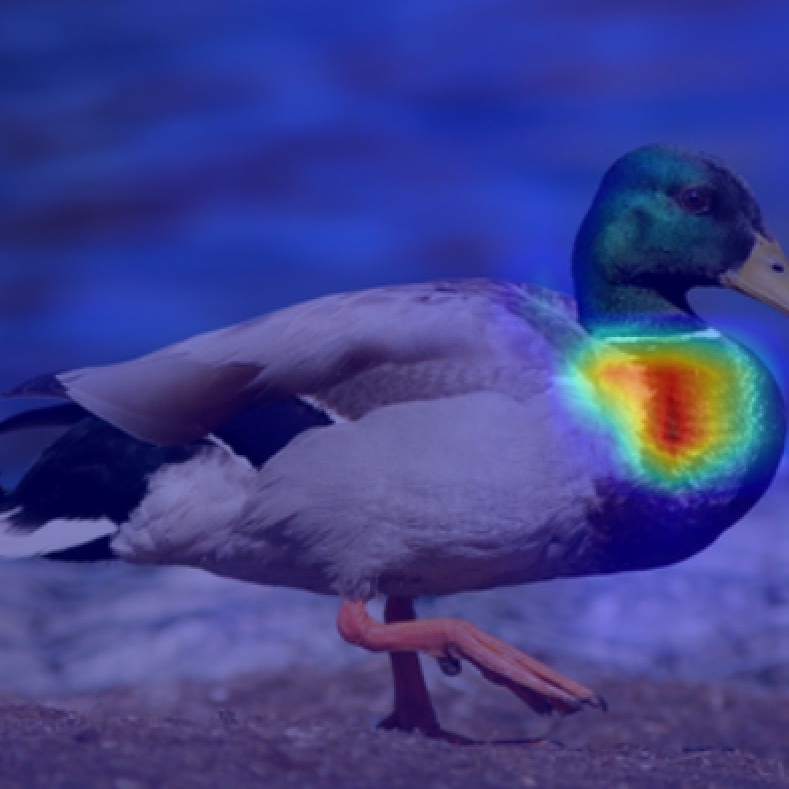}\vspace{1mm}
\includegraphics[width=1\linewidth]{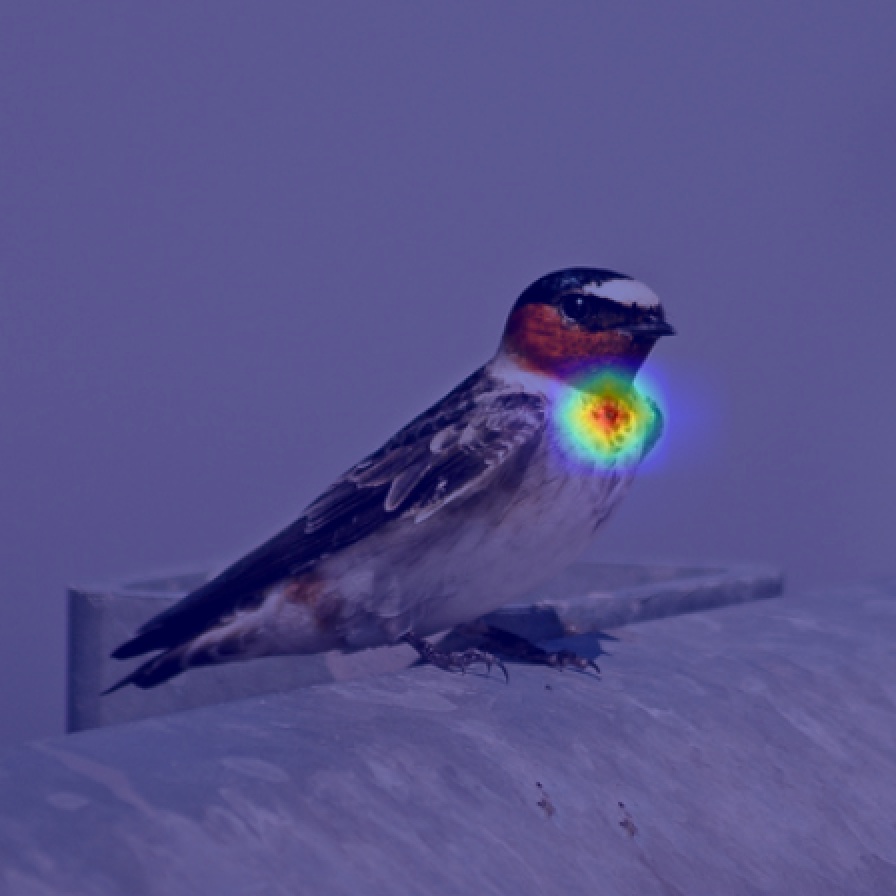}\vspace{1mm}
\includegraphics[width=1\linewidth]{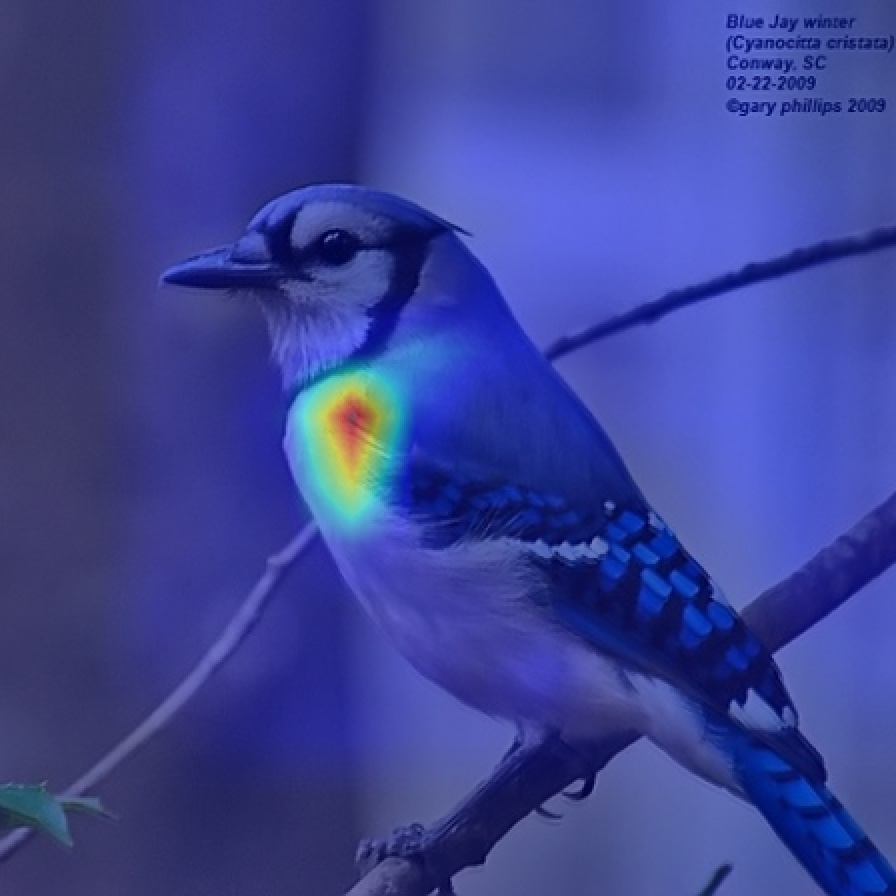}
\label{figure:breast}
\end{minipage}
}%
\subfigure[belly]{
\begin{minipage}[t]{0.15\linewidth}
\centering
\includegraphics[width=1\linewidth]{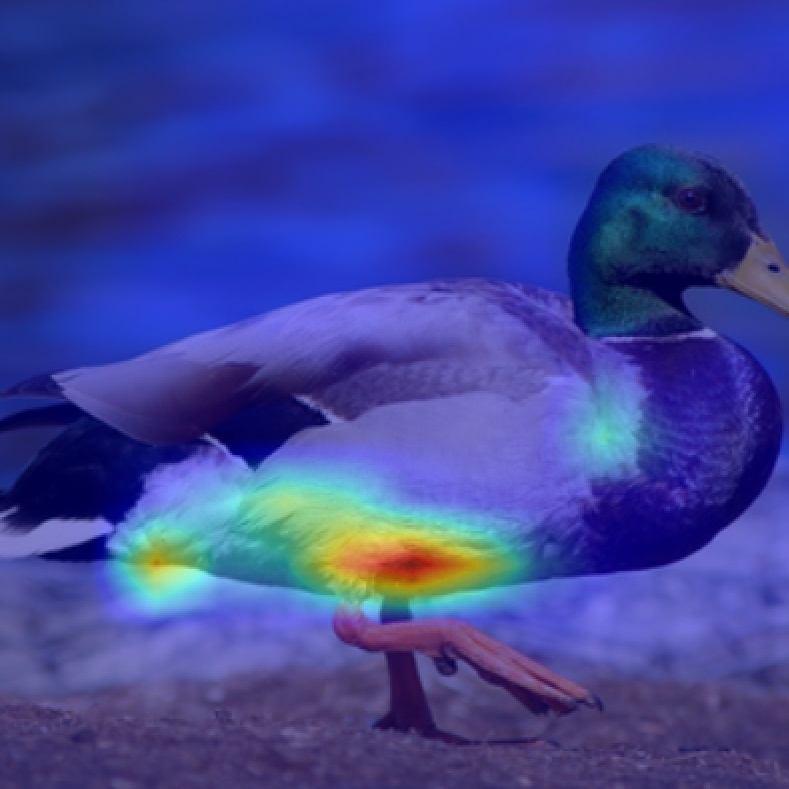}\vspace{1mm}
\includegraphics[width=1\linewidth]{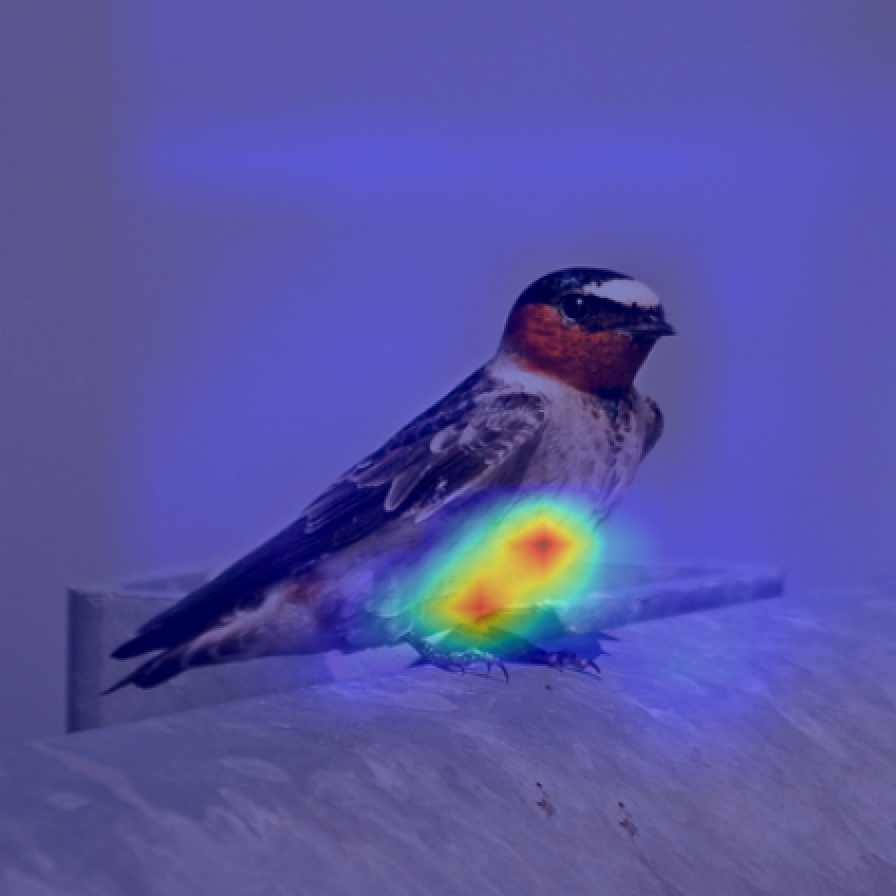}\vspace{1mm}
\includegraphics[width=1\linewidth]{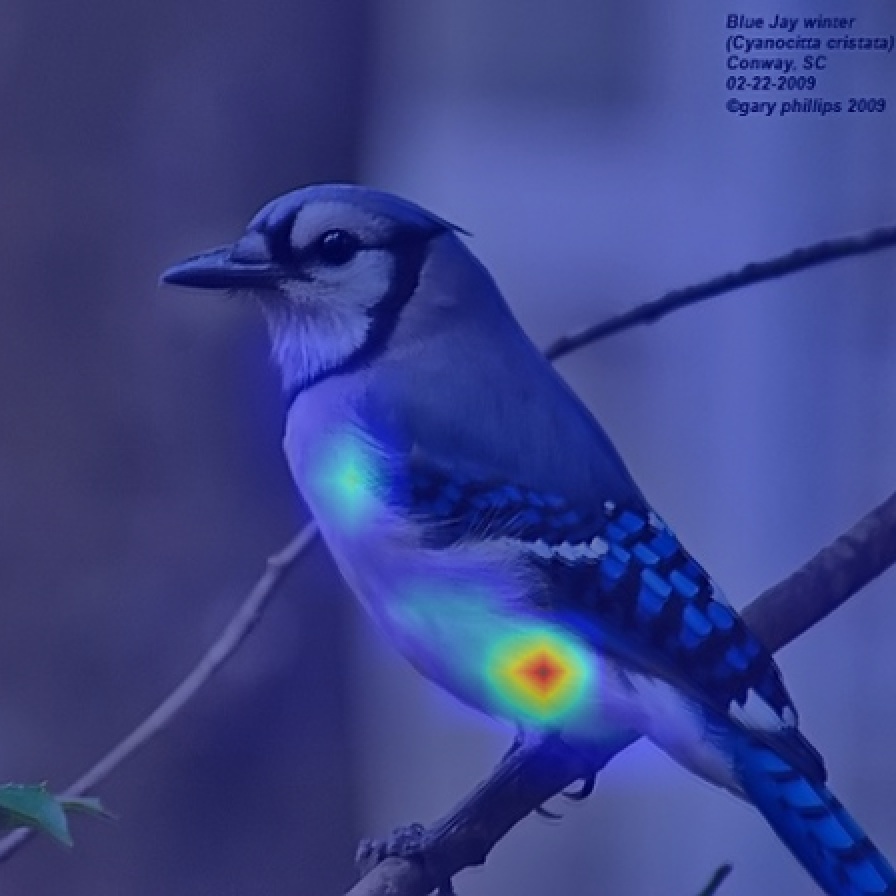}
\label{figure:belly}
\end{minipage}
}%
\subfigure[wing]{
\begin{minipage}[t]{0.15\linewidth}
\centering
\includegraphics[width=1\linewidth]{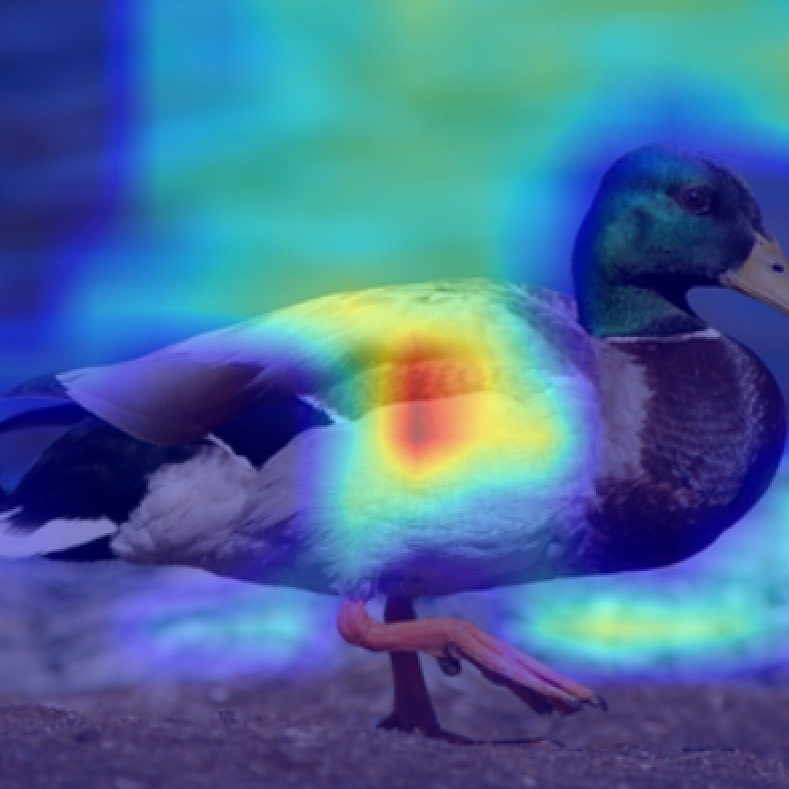}\vspace{1mm}
\includegraphics[width=1\linewidth]{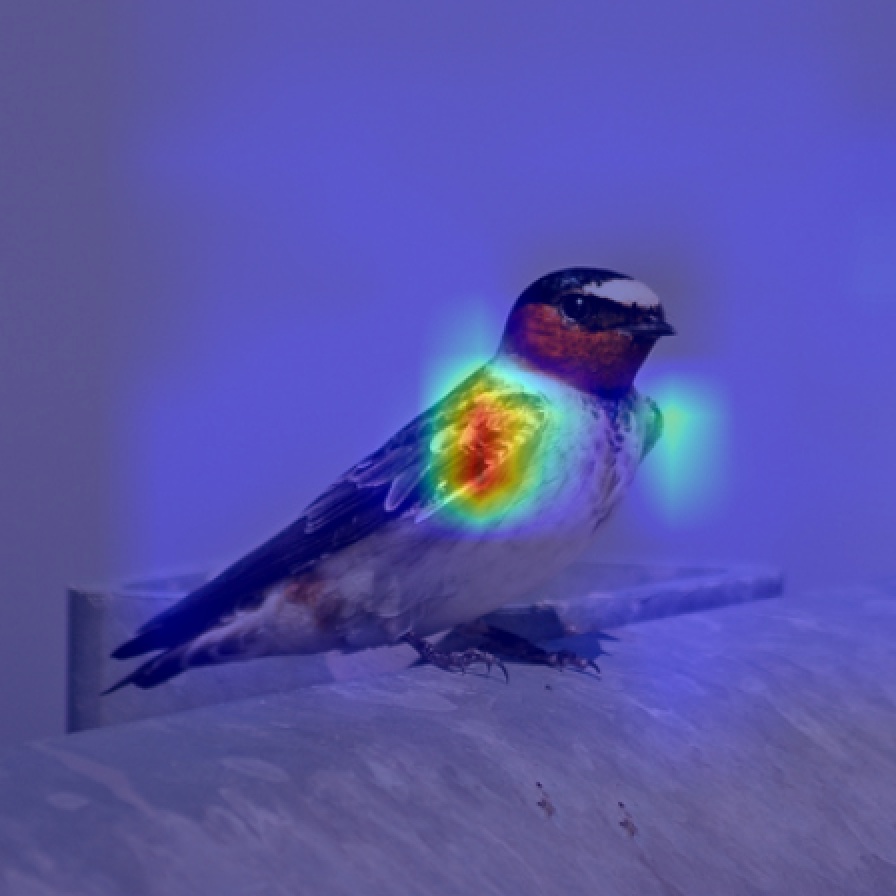}\vspace{1mm}
\includegraphics[width=1\linewidth]{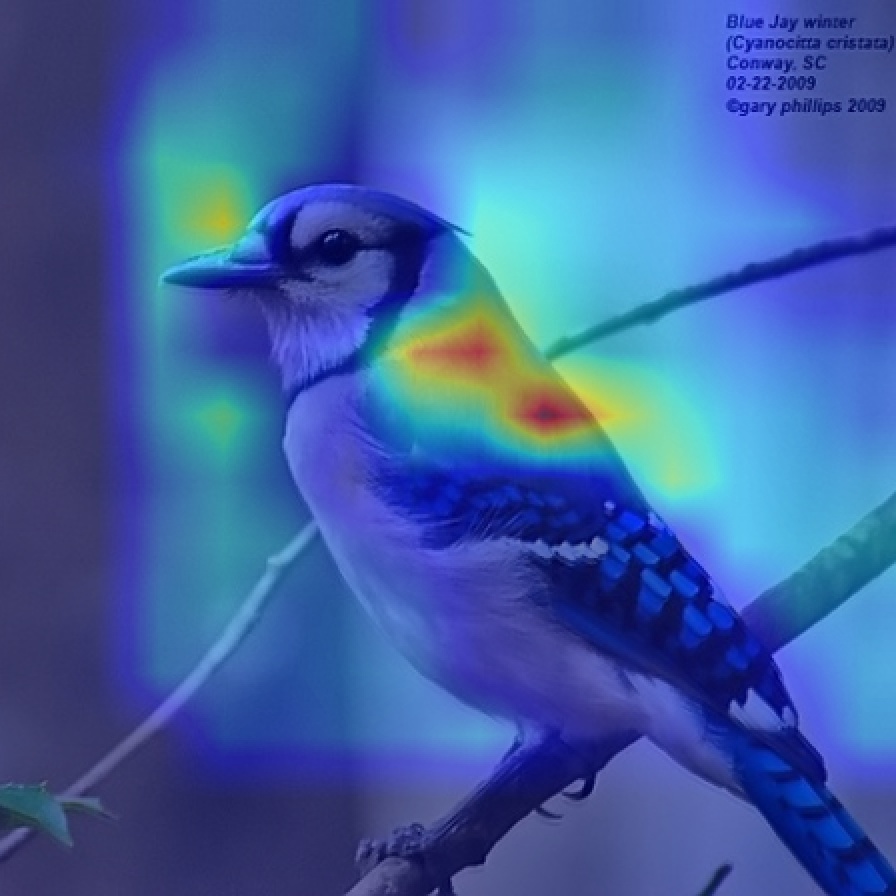}
\label{figure:wing}
\end{minipage}
}%
\centering
\vspace{-10pt}
\caption{Results of body part localization. Here each sample is selected from one of the three categories: "Mallard", "Cliff swallow" and "Blue jay". Then we pick five main parts of an bird body: eye, nape, breast, belly and wing. The results show that our method is capable of localizing those body parts accurately for the four samples.}
\vspace{-10pt}
\label{fig:localization}
\end{figure*}

\subsection{Discriminability Evaluation}
Fine-grained Recognition branch enables RSAN to discern fine-grained differences between similar classes. We in depth demonstrate this advantage by showing how our model outperforms others' by reducing misclassifications to similar classes. Specifically, in AWA2, among ten categories from the unseen domain according to the proposed splits \cite{XLSA18}, we further expect samples from unseen category "bobcat" would not be misclassified into those from similar seen categories like "leopard', "lion" in Figure \ref{fig:similar}. To demonstrate, here we choose two models GEM \cite{Liu2021GoalOrientedGE}, APN \cite{Xu2020AttributePN} for comparison, and for all the samples from an unseen category, we compute the proportion that they are classified into their own category and the similar seen categories. Figure \ref{fig:compare} shows the results. Our method (RSAN) achieves the largest proportion on "bobcat" while suppressing the proportion on other seen categories. GEM yields similar but less sharper results. APN not only fails to distinguish between its own category and three similar categories, it misclassifies "bobcat" into other dissimilar categories.

\subsection{Visualization of Attribute Localization}  
As stated in Section 3.3, our method is capable of localizing semantic attributes through attribute saliency maps. Now we demonstrate the locality of our method via overlapping the original image with the min-max normalized saliency map. Figure \ref{fig:localization} shows the results of three samples. Since in CUB dataset, attributes are organized in groups (e.g., a group of attributes describe the same part of a bird body but from different perspectives), we show the effect of body part localization for demonstration. Apparently, RSAN could not only localize the fine-grain attributes like "eye", but only localize the discriminative part of the coarse-grain attribute like "wing". Compared to the previous works \cite{Xu2020AttributePN, Liu2021GoalOrientedGE}, RSAN achieves a remarkable improvement on localization ability as each body part has been localize to a precise small region of the image. Meanwhile, we also observe that our model may have some imperfects in specific situations. For instance, our model localizes two discrete discriminative regions for the belly of "Cliff swallow" in Figure \ref{figure:belly}.

\section{Conclusion}
In this paper, we propose a novel ZSL framework named Region Semantically Aligned Network (RSAN), which transfers region-attribute alignment from seen classes to unseen classes. Specifically, Fine-grained Recognition branch is developed to obtain each attribute from a specific region of the image sample and exploit these attributes for recognition. Besides, Attribute Constraint branch is employed to regularize the image encoder shared with Fine-grained Recognition branch to extract robust and attribute-related visual features through attribute regressions with semantic knowledge. Experiments on several standard ZSL datasets reveal the benefit of our RSAN method, outperforming state-of-the-art methods. Besides, further experiments prove our model's locality of attributes and discriminability in face of similar classes. 

\begin{acks}
This work was supported in part by the National Natural Science Foundation of China under Grant 61806039 and 62073059, and in part by Sichuan Science and Technology Program under Grant 2020YFG0080 and 2020YFG0481. 
\end{acks}

\bibliographystyle{ACM-Reference-Format}
\bibliography{sample-base}










\end{document}